\documentclass{article}

\usepackage[main, final]{neurips_2026}

\usepackage[utf8]{inputenc} % allow utf-8 input
\usepackage[T1]{fontenc}    % use 8-bit T1 fonts
\usepackage{hyperref}       % hyperlinks
\usepackage{url}            % simple URL typesetting
\usepackage{booktabs}       % professional-quality tables
\usepackage{amsfonts}       % blackboard math symbols
\usepackage{nicefrac}       % compact symbols for 1/2, etc.
\usepackage{microtype}      % microtypography
\usepackage{xcolor}         % colors

\usepackage{graphicx}      % For \includegraphics
\usepackage{amsmath}       % For aligned environment and \text
\usepackage{amsfonts}      % Already there, good
\usepackage{amssymb}       % Additional math symbols
\usepackage{array}         % For table columns
\usepackage{wrapfig}  % 导言区添加

\usepackage{algorithm}
\usepackage{algpseudocode}
\usepackage{amsmath}
\usepackage{multirow}
\usepackage{marvosym}
\title{Transferable Multiform Attack for Cross-Modal Person Re-Identification}

\author{%
	Yunpeng Gong\textsuperscript{1},
	Can Yang\textsuperscript{1},
	Qingyuan Zeng\textsuperscript{2},
	Dejun Xu\textsuperscript{1},
	Zhiming Luo\textsuperscript{1}, \\
	{\bfseries Zhenzhong Wang\textsuperscript{1},
		Min Jiang\textsuperscript{1}\,\Letter} \\
	\textsuperscript{1}School of Informatics, Xiamen University \\
	\textsuperscript{2}The Hong Kong University of Science and Technology (Guangzhou)
}

\begin{document}

	\maketitle
	
	% Author contact information (camera-ready footnote)
	\begingroup
	\renewcommand{\thefootnote}{\fnsymbol{footnote}}
	\footnotetext[1]{Min Jiang (minjiang@xmu.edu.cn), Yunpeng Gong ( fmonkey625@gmail.com), Can Yang, Zhiming Luo and Zhenzhong Wang are with the Department of Artificial Intelligence, Key Laboratory of Multimedia Trusted Perception and Efficient Computing, Ministry of Education of China, School of Informatics, Key Laboratory of Digital Protection and Intelligent Processing of Intangible Cultural Heritage of Fujian and Taiwan, Ministry of Culture and Tourism, Xiamen University, Xiamen 361005, Fujian, P.R. China.\\
		\hspace*{2em} Corresponding author: Min Jiang.}
	\endgroup
	
\begin{abstract}
	Cross-modal person re-identification (ReID) adversarial attacks face a transferability challenge: existing methods may exhibit source-pair overfitting, where learned perturbations become overly dependent on the feature statistics of the source modality pair, thereby limiting their transferability under heterogeneous domain shifts. We argue that a key challenge is reducing source-pair-specific bias while maintaining attack effectiveness. To address this, we propose Multiform Attack (MA). The framework first learns a universal perturbation via Mahalanobis-guided gradient optimization. Since this perturbation may still retain source-pair-specific bias, the second stage performs sparse refinement through attention-shift-filtered multi-objective evolutionary search, leveraging heterogeneous source distributions in transfer settings. By separating universal perturbation learning from sparse evolutionary refinement, MA aims to reduce reliance on source-pair-specific patterns and improve transferability to unseen domains. Extensive experiments demonstrate strong attack effectiveness on cross-modal ReID benchmarks and improved transferability in the evaluated held-out-domain settings, with additional evaluations across model architectures.
\end{abstract}

	\section{Introduction}
	
	Person re-identification (ReID)~\cite{li2025video,zuo2024plip,zhang2024magic,yang2024robust,CFine,9018132,wu2024text} aims to match individuals across different cameras and time intervals. ReID tasks can be broadly divided into single-modality ReID~\cite{yang2023good,zuo2025cross,zhao2024unifying} and cross-modal ReID~\cite{gong2026theory,pang2024inter,shi2023dual,yang2022learning,Yang_2023_ICCV,lu2020cross}. The former matches query and gallery images within the same modality (e.g., RGB-RGB), whereas the latter performs matching across different modalities, including visible, infrared, thermal, and sketch images. Such cross-modal matching is particularly valuable in surveillance systems equipped with heterogeneous imaging devices, especially under varying illumination and environmental conditions.

\begin{figure}[t]   % 或 [t] / [b] / [htbp]
		\centering
		\includegraphics[width=1\linewidth]{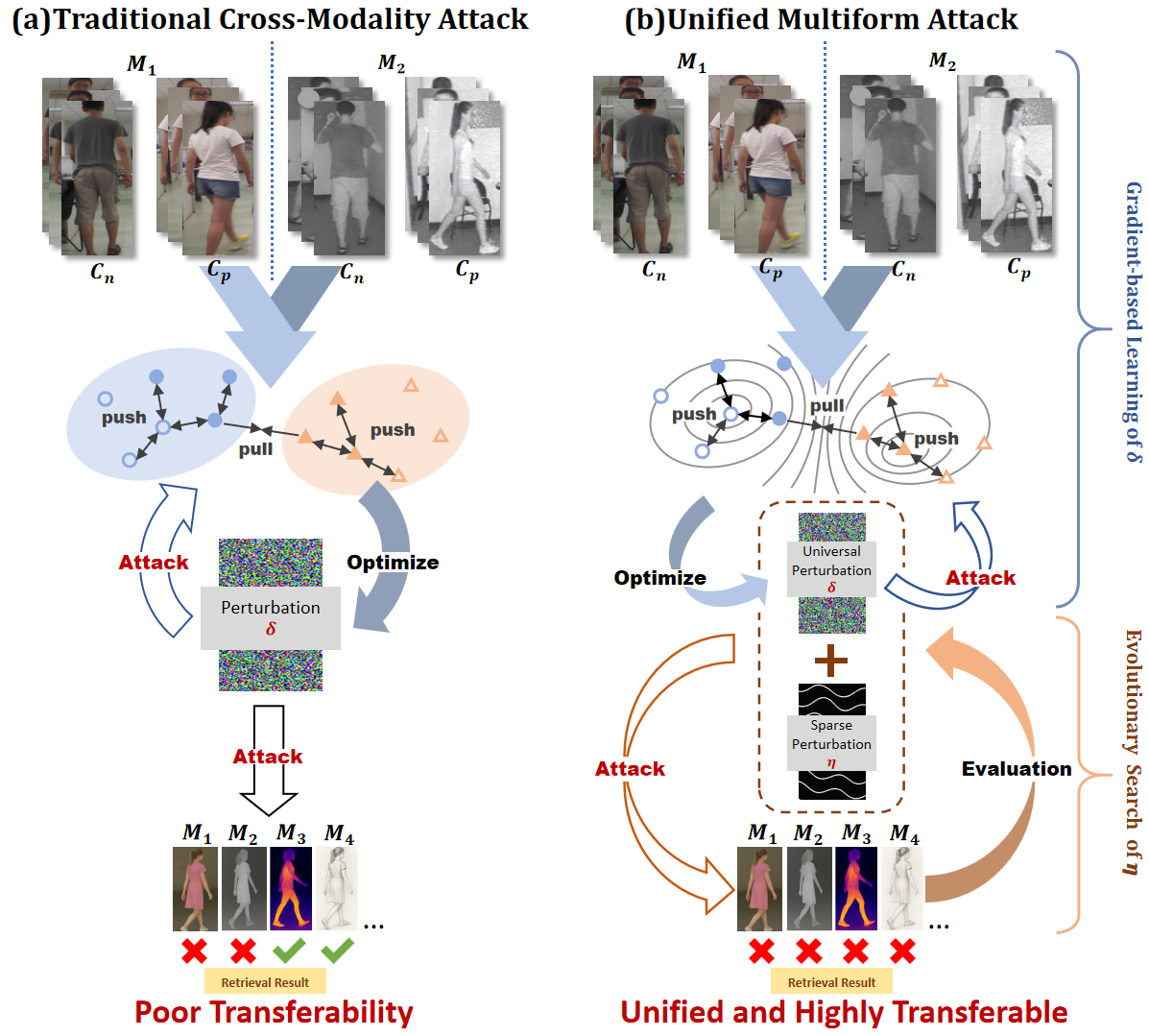}   % 宽度可按需调整
		\caption{Comparison between traditional gradient-based cross-modal attacks and the proposed MA. Conventional methods optimize perturbations over fixed source modality pairs, which may cause attacks to overfit source-pair-specific embedding relationships and limit their transferability to unseen modalities. MA couples Mahalanobis-guided universal perturbation learning with structure-aware sparse evolutionary refinement, shifting the optimization focus from source-pair effectiveness toward cross-modal transferability.}
		\label{pic1}
	\end{figure}

	With the widespread application of ReID in surveillance, criminal investigation, and intelligent security, concerns about its robustness to adversarial attacks have become increasingly prominent~\cite{zheng2023u,gong2024cross}. Security vulnerabilities in ReID systems can lead to serious consequences. For example, in surveillance systems used by law enforcement agencies, malicious insiders may exploit adversarial perturbations to mislead the ReID system, allowing criminals to evade identification. Existing adversarial attacks on ReID can be broadly classified into single-modality attacks~\cite{zheng2023u,gong2022person}, which operate within the same modality, and cross-modal attacks~\cite{gong2024cross}, which aim to transfer perturbations across different modality domains. Fig.~\ref{pic1} provides a comparison between traditional methods and our proposed approach. Despite significant progress, transferable adversarial attacks in cross-modal ReID remain underexplored. Single-modality attacks typically generalize poorly when applied to heterogeneous modalities, while existing cross-modal attack methods often optimize perturbations over fixed modality pairs, such as visible-infrared pairs. As a result, the learned perturbations may overfit to source-pair-specific embedding relationships, including distance structures and cross-modal alignment patterns, thereby limiting their transferability to unseen modalities.
	
	%\begin{wrapfigure}{r}{0.5\linewidth}
	%	\centering
	%	\includegraphics[width=1\linewidth]{pipline6.png}
	%	\caption{\small Comparison between traditional gradient-based cross-modal attacks and the proposed MA. Conventional methods optimize perturbations over fixed source modality pairs, which may cause attacks to overfit source-pair-specific embedding relationships and limit their transferability to unseen modalities. MA couples Mahalanobis-guided universal perturbation learning with structure-aware sparse evolutionary refinement, progressively improving the perturbation from source-pair effectiveness toward cross-modal generalization.}
	%	\label{pic1}
	%\end{wrapfigure}
	
	To address these limitations, we focus on improving the transferability of cross-modal adversarial attacks and aim to construct a generalizable attack framework that remains effective across heterogeneous modality domains. However, this goal remains challenging due to three critical obstacles. First, modality-specific overfitting causes adversarial perturbations optimized on source domains to fail when transferred to unseen modalities. Second, source-pair feature bias makes perturbations optimized on fixed modality pairs capture source-pair-specific embedding relationships rather than transferable attack patterns. Third, optimization trade-offs arise because improving attack effectiveness for one modality pair may weaken its generalization to other modalities.
	
	The key idea of MA is that transferable attacks across heterogeneous modalities can be viewed as a generalization-oriented optimization problem. A perturbation should be sufficiently effective on accessible source modality pairs, while avoiding overfitting to source-pair-specific embedding relationships. Existing cross-modal attacks usually optimize a single perturbation over fixed modality pairs, implicitly assuming that one perturbation can simultaneously achieve strong source-pair attack effectiveness and cross-modal generalization. However, when the modality gap is large, this assumption becomes restrictive: the optimized perturbation may rely on the distance relationships and alignment patterns specific to the source modality pair, thereby limiting its transferability to unseen modalities.
	
	\begin{figure*}[t]
		\centering
		\includegraphics[width=1\linewidth]{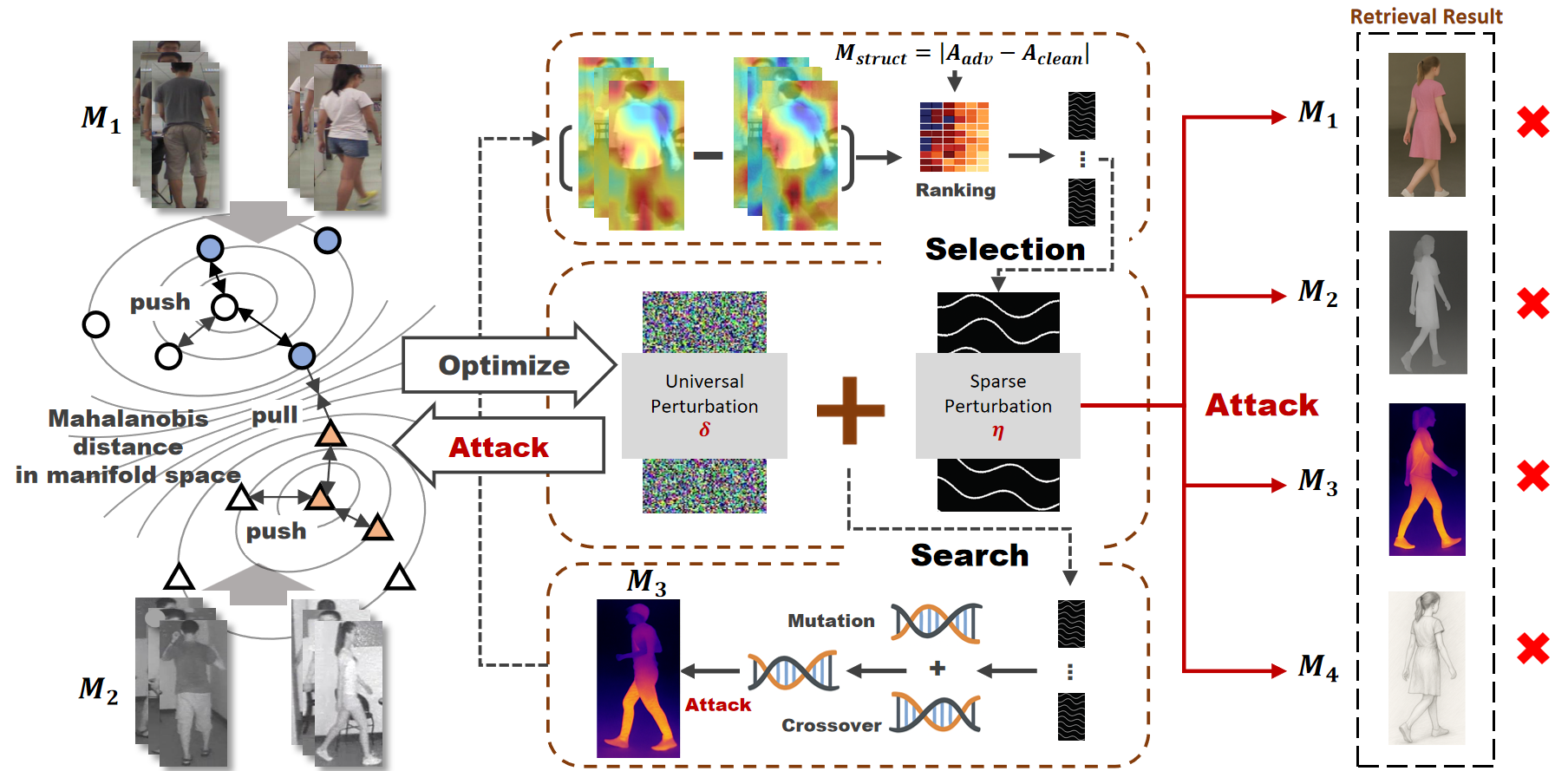}
		\caption{Overview of the proposed cross-modal transferable adversarial attack framework. A universal perturbation is initially learned via Mahalanobis-guided gradient optimization, providing a strong prior. Subsequently, a structure-aware sparse evolutionary search is performed, where an attention-shift filter retains candidates exhibiting the largest measured attention-map shifts relative to clean inputs. The search optimizes sparse pixel-level corrections to the fixed universal perturbation while respecting the overall perturbation budget. The final adversarial example, combining both components, is evaluated for its ability to degrade retrieval performance across heterogeneous modality models.}
		\label{method}
	\end{figure*}

	Motivated by this observation, we propose MA, a novel adversarial attack framework for transferable cross-modal ReID attacks. MA decomposes transferable attack generation into two complementary processes. The first stage learns a source-optimized universal perturbation on source modality pairs through Mahalanobis-guided gradient optimization, providing an effective attack direction for subsequent optimization. Nevertheless, since this perturbation is still driven by the gradients and embedding structure of the source pairs, it may retain source-pair-specific bias. The second stage performs sparse evolutionary refinement, not to simply add modality-specific patches, but to make constrained attention-guided sparse adjustments to the first-stage perturbation, with the aim of reducing reliance on source-pair-specific patterns.
	Evolutionary search is adopted in the second stage instead of gradient descent because multi-modal sequential gradient optimization suffers from a dilemma: further fine-tuning on heterogeneous models can substantially alter the universal attack direction learned in the first stage, while introducing new modality-specific biases that again entangle the perturbation with particular source distributions. By contrast, the universal perturbation provides a high-quality prior that restricts the search space to sparse adjustments around the learned attack direction. Within this space, evolutionary search seeks to reduce reliance on any single accessible source distribution by maintaining solution diversity and imposing selection pressure toward cross-modal generalization, avoiding further sequential gradient updates that may introduce modality-specific bias.
	As shown in Fig.\ref{method}, MA implements this two-stage design through Mahalanobis-guided universal perturbation learning followed by structure-aware sparse evolutionary refinement. The pipeline shifts its optimization focus from source-pair effectiveness toward cross-modal transferability.

	\textbf{Our main contributions are summarized as follows:}
	\begin{itemize}
		\item Transferable cross-modal adversarial attack is studied from a generalization perspective, and MA is proposed to decompose it into universal perturbation learning and generalization-oriented sparse refinement.
		\item A two-stage optimization strategy is introduced, combining Mahalanobis-guided universal perturbation learning with structure-aware sparse evolutionary refinement; a theoretical stability justification is provided in the supplementary material.
		\item Extensive experiments demonstrate superior cross-modal transferability across diverse datasets, models, and modality settings.
	\end{itemize}

	\section{Related Work}
	
	\textbf{Adversarial attacks on ReID.}
	Adversarial attacks for person re-identification have been predominantly studied under single-modality visible-visible matching. 
	Early efforts extended classification-based attacks to the retrieval setting: Metric-FGSM~\cite{bai2020adversarial} adapts FGSM~\cite{fgsm} by perturbing selected samples to manipulate feature distances, but its reliance on per-sample selection incurs high computational cost. 
	To alleviate this dependency, Gong et al.~\cite{gong2022person} introduced a color-based attack that operates without reference images, and ODFA~\cite{zheng2023u} further improved semantic disruption by perturbing feature representations along opposite directions. 
	Building on the idea of universal perturbations, Col.+Del.~\cite{yang2023towards} combined UAP-Retrieval~\cite{li2019universal} with color-space perturbations~\cite{laidlaw2019functional} to enhance attack generality. 
	Despite these advances, all the above methods are designed for RGB-based matching and do not explicitly address the feature distribution gap across heterogeneous modalities.
	
	Recent work has begun to investigate cross-modal ReID attacks. 
	CMPS~\cite{gong2024cross} jointly optimizes shared perturbations by aggregating gradients across visible-infrared modality pairs, improving cross-modal transferability over single-modality methods. 
	However, its optimization remains coupled to the specific modality pair seen during training, making the learned perturbation prone to overfitting to source-pair-specific embedding structures and scaling poorly to unseen modality combinations. 
	To improve scalability, MUA~\cite{bian2025modality} assigns a dedicated generator to each known modality, but this per-modality design requires separate generator training for each accessible modality. 
	In contrast, our method decouples perturbation generation from fixed source modality pairs and aims to learn perturbations that generalize to heterogeneous and previously unseen modality settings.
	
	\textbf{Sparse and evolutionary adversarial attacks.}
	Beyond gradient-based methods, evolutionary algorithms have been explored for generating sparse adversarial perturbations in discrete pixel spaces where gradient signals are unreliable~\cite{su2019one}.
	Subsequent work~\cite{Croce_2019_ICCV,dong2020greedyfool,williams2023black,NEURIPS2023_d482f136} improved attack success under restricted perturbation budgets, demonstrating that sparse perturbations can be highly effective while remaining imperceptible.
	However, these methods are designed for classification tasks and optimize perturbations against a single target model, without addressing the cross-model or cross-modal generalization required for ReID attacks.
	In contrast, our evolutionary refinement stage is specifically designed for cross-modal transferability: rather than searching from scratch, it searches for sparse corrections around a learned universal perturbation and leverages attention-shift filtering to identify structurally impactful sparse corrections, guiding the search toward perturbations that remain effective across heterogeneous models.

	\section{Methodology}

We aim to construct an adversarial perturbation that remains effective on accessible source modality pairs while generalizing to unseen target modalities. 
Given a query image \(x\), we seek a perturbation \(\Delta = \delta + \eta\) consisting of two components: a universal perturbation \(\delta\) that captures an attack direction shared among accessible source modalities, and a sparse correction \(\eta\) that aims to mitigate source-pair-specific overfitting to improve cross-modal transferability. 
The perturbed query \(x_{\mathrm{adv}} = x + \Delta\) is then used to mislead the victim re-identification model \(\mathcal{M}\) when retrieving from a clean gallery.
The complete optimization procedure is detailed in Algorithm~1 (supplementary material). We provide a theoretical analysis of the evolutionary refinement stage in the supplementary material.

\subsection{Mahalanobis-Guided Universal Perturbation Learning}

In the first stage, we learn a universal perturbation on accessible source modality pairs using gradient-based optimization. Given a mini-batch of source query images $\{x_i\}_{i=1}^{n}$, the objective is formulated as
\begin{equation}
	\mathcal{L}(\delta)= \frac{1}{n} \sum_{i=1}^{n} \mathcal{L}_{\mathrm{tri}}(\delta, x_i),
\end{equation}
where $\delta$ denotes the universal perturbation and $\mathcal{L}_{\mathrm{tri}}$ is an attack-oriented triplet loss defined in the feature space of the ReID model.

Most existing ReID attack methods~\cite{gong2024cross,zheng2023u,yang2023towards,gong2022person,bai2020adversarial} use Euclidean distance to construct metric-learning losses. However, Euclidean distance treats all feature dimensions equally and ignores the anisotropic covariance structure of ReID embeddings. To incorporate such feature-space statistics, we use the Mahalanobis distance computed from accessible source training features. This distance reweights feature directions according to their covariance and therefore provides a covariance-aware metric based on source feature statistics for perturbation optimization.

Let
\begin{equation}
	f_i = F(x_i), \qquad f_i^{\mathrm{adv}} = F(\mathrm{clip}(x_i+\delta)),
\end{equation}
where $F(\cdot)$ denotes the feature extractor of the source ReID model. For each source modality $m \in \{m_1,m_2\}$, we precompute clean feature centroids from the source training split. For image $x_i$, $C^{m}_{p}(x_i)$ denotes the nearest positive centroid associated with the same identity, and $C^{m}_{n}(x_i)$ denotes the selected negative centroid associated with a different identity. The attack-oriented triplet loss is defined as
\begin{equation}
	\begin{aligned}
		\mathcal{L}_{\mathrm{tri}}(\delta, x_i)
		&= \left[
		D_M\!\left(C^{m_1}_{n}(x_i), f_i^{\mathrm{adv}}\right)
		-
		D_M\!\left(C^{m_1}_{p}(x_i), f_i^{\mathrm{adv}}\right)
		+ \rho
		\right]_+ \\
		&\quad +
		\left[
		D_M\!\left(C^{m_2}_{n}(x_i), f_i^{\mathrm{adv}}\right)
		-
		D_M\!\left(C^{m_2}_{p}(x_i), f_i^{\mathrm{adv}}\right)
		+ \rho
		\right]_+ .
	\end{aligned}
\end{equation}
Minimizing this loss encourages the adversarial feature to move closer to negative centroids while moving away from positive centroids, thereby disrupting the correct query–gallery matching relationship. The margin $\rho > 0$ controls the desired separation, and $[\cdot]_+ = \max(0,\cdot)$ denotes the hinge function.

For feature vectors $u$ and $v$, the squared Mahalanobis distance is defined as
\begin{equation}
	D_M(u,v) = (u-v)^\top \Sigma_{\lambda}^{-1}(u-v),
\end{equation}
where $\Sigma_{\lambda}$ is the regularized feature covariance matrix estimated from clean source training features:
\begin{equation}
	\Sigma_{\lambda}
	=
	\frac{1}{N-1}\sum_{j=1}^{N}
	(F(x_j)-\mu)(F(x_j)-\mu)^\top
	+ \lambda I ,
	\qquad
	\mu = \frac{1}{N}\sum_{j=1}^{N} F(x_j).
\end{equation}
Here, $\lambda I$ is a diagonal regularizer used to ensure numerical stability when inverting the covariance matrix. All covariance statistics and centroids are computed only from accessible source training data, without using target-domain test samples.

At iteration $t$, the accumulated attack-objective gradient is updated as
\begin{equation}
	v_{t+1}
	=
	\beta v_t
	+
	(1-\beta)
	\cdot
	\frac{-\nabla_{\delta}\mathcal{L}(\delta_t)}
	{\|\nabla_{\delta}\mathcal{L}(\delta_t)\|_1},
\end{equation}
where $v_t$ is the exponential moving average of the attack-objective gradient and $\beta$ is the momentum decay coefficient. The universal perturbation is then updated by
\begin{equation}
	\delta_{t+1}
	=
	\Pi_{\|\delta\|_{\infty}\leq \varepsilon}
	\left(
	\delta_t + \alpha \cdot \mathrm{sign}(v_{t+1})
	\right),
\end{equation}
where $\Pi_{\|\delta\|_{\infty}\leq \varepsilon}(\cdot)$ denotes projection onto the $\ell_{\infty}$ perturbation ball. In our experiments, we set $\beta=0.9$, $\alpha=\varepsilon/10$, and $\varepsilon=8$ (on the raw pixel scale from 0 to 255) unless otherwise specified.

This stage produces a strong universal attack direction by optimizing a covariance-aware metric loss on accessible source modality pairs. The resulting perturbation serves as the base component for the subsequent sparse evolutionary refinement stage.

\subsection{Evolutionary Search} 

In the second optimization stage, we employ evolutionary search to improve the cross-modal transferability of perturbations.  
Specifically, evolutionary algorithms are used to identify sparse corrections $\eta$ that, when combined with the universal perturbation $\delta$ from the first stage, further enhance adversarial performance across the evaluated modality settings.  
The optimization objective is: 
\begin{equation} 
	\min\nolimits_{\eta}\; \mathit{\Phi}\bigl(\mathcal{F}(x_{\mathrm{adv}})\bigr) 
	\quad \text{s.t.} \quad 
	\| \eta \|_{0,\mathrm{sp}} \leq k,\; 
	\| \delta + \eta \|_{\infty} \leq \varepsilon,\; 
	\eta[p,c] \in \{-1,0,+1\}. 
\end{equation} 
where \(k\) represents the maximum number of perturbed pixel locations in the sparse refinement, and \(\eta[p,c]\) denotes the correction value at pixel location \(p\) and channel \(c\).  
The spatial count \(\|\eta\|_{0,\mathrm{sp}}\) counts a location once if any channel is nonzero.  
In practice, after initialization, crossover, and mutation, we apply a repair operation to keep at most \(k\) perturbed pixel locations, set all other entries to zero, and ensure that each correction value remains in \(\{-1,0,+1\}\).  
If a correction value would violate the global perturbation constraint, i.e., 
$|\delta[p,c]+\eta[p,c]|>\varepsilon$, it is removed.  
Therefore, every candidate satisfies 
\(\|\eta\|_{0,\mathrm{sp}} \leq k\) and 
\(\|\delta+\eta\|_\infty \leq \varepsilon\) before fitness evaluation. 

The adversarial example is constructed as 
\(x_{\mathrm{adv}} = x + \delta + \eta\), 
with its feature representation 
\(f_{\mathrm{adv}} = \mathcal{F}(x_{\mathrm{adv}})\).  
The multi-objective vector is: 
\begin{equation} 
	\mathit{\Phi}(f_{\mathrm{adv}}) =  
	\left( 
	\tilde{\mathcal{D}}(f_{\mathrm{adv}}),\  
	\tilde{\mathcal{S}}(f_{\mathrm{adv}}),\  
	\|\eta\|_2 
	\right)^\top . 
\end{equation} 
The sparse correction is searched in a discrete space.  
Specifically, each candidate $\eta$ contains at most $k$ perturbed pixel locations, and the channel-wise correction values at selected locations are restricted to $\{-1,0,+1\}$.  
The inclusion of zero values naturally supports sparse perturbation patterns, while minimizing $\|\eta\|_2$ further encourages compact sparse corrections.  
$\mathcal{D}(f_{\mathrm{adv}})$ and $\mathcal{S}(f_{\mathrm{adv}})$ denote the raw metric objective and the retrieval attack success rate across all surrogate models, respectively.  
Their transformed forms, $\tilde{\mathcal{D}}(f_{\mathrm{adv}})$ and $\tilde{\mathcal{S}}(f_{\mathrm{adv}})$, are used as minimization objectives.  

For model $\mathcal{M}_i$, the metric objective is defined as: 
\begin{equation} 
	\mathcal{D}_i(f_{\mathrm{adv}}) =  
	(f_{\mathrm{adv}} - C_i)^\top S_i^{-1} (f_{\mathrm{adv}} - C_i), 
\end{equation} 
where $\mathcal{D}_i(f_{\mathrm{adv}})$ denotes the metric deviation induced by the perturbed input on model $\mathcal{M}_i$, measured using the squared Mahalanobis distance to the corresponding cluster centroid $C_i$.  
Here, $S_i$ denotes the covariance matrix estimated in the feature space of model $\mathcal{M}_i$.  
The total metric objective across all surrogate models is: 
\begin{equation} 
	\mathcal{D}(f_{\mathrm{adv}}) = \sum_{i=1}^{n} \mathcal{D}_i(f_{\mathrm{adv}}), 
\end{equation} 
where $n$ is the total number of surrogate models involved in the evolutionary search.  
To convert it into a minimization objective, we use: 
\begin{equation} 
	\tilde{\mathcal{D}}(f_{\mathrm{adv}})=  
	\exp\left(-\mathcal{D}(f_{\mathrm{adv}})\right). 
\end{equation} 
For model $\mathcal{M}_i$, we define the retrieval-based attack success indicator under the standard ReID inference protocol as: 
\begin{equation} 
	\mathcal{S}_i(f_{\mathrm{adv}}) =  
	\begin{cases}  
		1, & \text{if } \mathrm{id}\bigl(g_i^{(1)}(x_{\mathrm{adv}})\bigr) \neq y,\\ 
		0, & \text{otherwise}, 
	\end{cases} 
\end{equation} 
where $g_i^{(1)}(x_{\mathrm{adv}})$ denotes the top-1 retrieved gallery image returned by model $\mathcal{M}_i$ for the adversarial query $x_{\mathrm{adv}}$, and $y$ is the ground-truth identity of the query.  
The overall retrieval attack success rate across surrogate models is: 
\begin{equation} 
	\mathcal{S}(f_{\mathrm{adv}}) =  
	\frac{1}{n} \sum_{i=1}^{n} \mathcal{S}_i(f_{\mathrm{adv}}). 
\end{equation} 
To convert it into a minimization objective, we define: 
\begin{equation} 
	\tilde{\mathcal{S}}(f_{\mathrm{adv}}) =  
	1 - \mathcal{S}(f_{\mathrm{adv}}). 
\end{equation}

\textbf{Structure-Aware Filtering.}  
Prior to fitness evaluation, we introduce a structure-aware guidance module that filters perturbation individuals based on attention shift metrics, before subsequent multi-objective evaluation.  
For each candidate $\eta$, we construct $x_{\mathrm{adv}}$ and extract its attention heatmap $A_{\mathrm{adv}}$ from the ReID model.  
We compare $A_{\mathrm{adv}}$ with the clean image's attention heatmap $A_{\mathrm{clean}}$ and compute the structure difference map: 
\begin{equation} 
	M_{\mathrm{struct}} = |A_{\mathrm{adv}} - A_{\mathrm{clean}}|. 
\end{equation} 
The $L_1$ norm $\|M_{\mathrm{struct}}\|_1$ quantifies the attention shift.  
Candidates are ranked by this score, and only the top-$K$ candidates with the largest measured attention shifts are retained for multi-objective fitness evaluation.

\textbf{Domination Definition.} 
During the selection phase, we employ Pareto-based non-dominated sorting to jointly optimize attack effectiveness and perturbation compactness.  
Given the fixed universal perturbation $\delta$, each candidate sparse correction $\eta$ is evaluated by the multi-objective vector 
$\Phi(\eta)$, which consists of the transformed metric objective, the transformed retrieval attack objective, and the correction compactness term.  
For two candidate corrections $\eta_i$ and $\eta_j$ with objective vectors 
$\Phi(\eta_i)$ and $\Phi(\eta_j)$,  
$\eta_i$ dominates $\eta_j$ if 
\begin{equation} 
	\Phi_m(\eta_i) \leq \Phi_m(\eta_j),\quad \forall m, 
	\quad \text{and} \quad 
	\exists m,\ \Phi_m(\eta_i) < \Phi_m(\eta_j). 
\end{equation} 
Here, a smaller $\tilde{\mathcal{S}}(f_{\mathrm{adv}})$ corresponds to a higher cross-model retrieval attack success rate, while $\|\eta\|_2$ encourages compact sparse corrections.  
Detailed selection, crossover, and mutation operations are provided in the supplementary materials.  
This two-stage framework first learns a universal perturbation $\delta$ via gradient-based optimization.  
After $\delta$ is learned, it is fixed as the base perturbation, and structure-aware evolutionary search is used to optimize only the sparse correction $\eta$ for enhancing cross-modal transferability.

	%\begin{table}[t]\scriptsize
	%	\centering
	%	\caption{Ablation studies on the DEEN baseline using the LLCM dataset. {Maha.} refers to Mahalanobis distance, {Grad.} represents Gradient-based optimization, {Evol.} indicates Evolutionary Search, and {Struct.} is the proposed Structure-Aware Filtering module.}
	%	\begin{tabular}{ccccc|cc}
		%		\toprule
		%		No. & Maha. & Grad. & Evol. & Struct. & mAP & rank-1 \\
		%		\midrule
		%		0 & $\times$ & $\times$ & $\times$ & $\times$ & 62.53 & 65.84 \\
		%		1 & $\times$ & $\checkmark$ & $\times$ & $\times$ & 8.54 & 12.35 \\
		%		2 & $\checkmark$ & $\checkmark$ & $\times$ & $\times$ & 6.88 & 9.47 \\
		%		3 & $\checkmark$ & $\checkmark$ & $\checkmark$ & $\times$ & 3.43 & 2.49 \\
		%		4 & $\checkmark$ & $\checkmark$ & $\checkmark$ & $\checkmark$ & \textbf{3.01} & \textbf{1.97} \\
		%		\bottomrule
		%	\end{tabular}
	%	\label{ablation_llcm}
	%\end{table}

	\section{Empirical Study}
	In all experiments, adversarial perturbations are constrained by the $\ell_\infty$ norm with budget $\varepsilon=8$. For single-channel modalities (e.g., thermal or infrared), images are replicated to three channels, consistent with the standard input format of ReID models. The non-visible to visible setting evaluates whether the learned perturbation transfers in the reverse direction, providing a complementary assessment of cross-modal transferability under the standard retrieval protocol.

	\subsection{Dataset and Evaluation Metric}
	We evaluate our method on five cross-modality ReID datasets, including four existing benchmarks, RegDB~\cite{nguyen2017person}, LLCM~\cite{zhang2023diverse}, SYSU~\cite{wu2017rgb}, and Sketch~\cite{pang2018cross}, as well as our constructed CnMix dataset. SYSU includes 395 training identities captured by six cameras (four Visible and two near-infrared), with 22,258 visible and 11,909 near-infrared images, and a test set of 95 identities with 3,803 query images. LLCM is designed for low-light cross-modality ReID, presenting diverse scenarios and challenges for adversarial attacks. Sketch dataset includes 200 individuals, each represented by one sketch and two photographs, captured by two cross-view cameras during daylight and manually cropped to focus on the individual. With increasingly stringent privacy regulations, personal data processing technologies face higher compliance requirements. To address this, some ReID systems utilize special image transformation methods or leverage multimodal data to enhance privacy protection~\cite{kieu2019domain,ouchi2023privacy}. We created CnMix by applying random channel mixing to Market1501~\cite{zheng2015scalable} images, simulating privacy-preserving image transformations (details in Supplementary Materials).
	Following~\cite{gong2024cross,zheng2015scalable}, we use Rank-k precision, Cumulative Matching Characteristics (CMC), and mean Average Precision (mAP) as evaluation metrics. Rank-1 measures the accuracy of the top result, while mAP evaluates the mean accuracy based on ranked query results. In adversarial attacks, lower accuracy scores indicate higher attack effectiveness.
	
	\begin{table*}[]\tiny
		\centering
		\caption{Results for attacking cross-modality ReID systems on the RegDB~\cite{nguyen2017person} dataset. It reports on visible images querying thermal images and vice versa. Rank at \( r \) accuracy (\%) and mAP (\%) are reported. The perturbation $\epsilon$ is set to 8.}
		\label{tab:table1}
		\resizebox{\textwidth}{!}{ % Scale the table to the width of the text block
			\begin{tabular}{llccccccccc}
				\toprule
				\multicolumn{2}{c}{Settings} & \multicolumn{4}{c}{Visible to Thermal} & \multicolumn{4}{c}{Thermal to Visible} \\
				\cmidrule(r){1-2} \cmidrule(lr){3-6} \cmidrule(l){7-10}
				Method & Venue & \( r = 1 \) & \( r = 10 \) & \( r = 20 \) & mAP & \( r = 1 \) & \( r = 10 \) & \( r = 20 \) & mAP \\
				\midrule
				AGW baseline~\cite{ye2021deep}  & TPAMI 2022 & 70.05 & 86.21 & 91.55 & 66.37 & 70.49 & 87.21 & 91.84 & 65.90 \\
				M-FGSM attack~\cite{bai2020adversarial} & TPAMI 2020 &29.34&52.90&61.44&23.35 &23.64&40.36&48.61&18.57 \\
				LTA attack~\cite{gong2022person} & CVPR 2022 & 12.65 & 25.24 & 34.02 & 12.80 & 10.51 & 22.93 & 31.79 & 9.74 \\
				ODFA attack~\cite{zheng2023u} & IJCV 2023 &28.57&51.42&60.58&21.84 &17.26&33.27&42.92&15.27 \\
				Col.+Del. attack~\cite{yang2023towards} & TPAMI 2023 &5.12&16.83&22.10&4.94&4.92&14.47&23.04&4.86 \\
				CMPS attack~\cite{gong2024cross} & NeurIPS 2024  & 2.29 & 9.06 & 18.35 & 3.92 & 1.93 & 11.44 & 19.30 & 3.46 \\
				MUA attack~\cite{bian2025modality} & TIFS 2025 & 1.85 & 7.84 & 16.95 & 3.21 & 1.52 & 10.31 & 18.02 & 2.98 \\
				Our attack & NeurIPS 2026 & \textbf{0.83} & \textbf{6.24} & \textbf{9.27} & \textbf{1.08} & \textbf{0.91} & \textbf{7.47} & \textbf{10.02} & \textbf{1.12} \\
				\bottomrule
			\end{tabular}
		}
		\label{regdb}
	\end{table*}

	\begin{table*}[h]\tiny
		\centering
		\caption{Results for attacking cross-modality ReID systems on the LLCM~\cite{zhang2023diverse} dataset. It reports on visible images querying Infrared images and vice versa. Rank accuracy (\%) and mAP (\%) are reported.}
		\label{tab:table1}
		\resizebox{\textwidth}{!}{ % Scale the table to the width of the text block
			\begin{tabular}{llccccccccc}
				\toprule
				\multicolumn{2}{c}{Settings} & \multicolumn{4}{c}{Visible to Infrared} & \multicolumn{4}{c}{Infrared to Visible} \\
				\cmidrule(r){1-2} \cmidrule(lr){3-6} \cmidrule(l){7-10}
				Method & Venue & \( r = 1 \) & \( r = 10 \) & \( r = 20 \) & mAP & \( r = 1 \) & \( r = 10 \) & \( r = 20 \) & mAP \\
				\midrule
				DEEN baseline~\cite{zhang2023diverse}  & CVPR 2023 &62.53  &90.31  &94.73  &65.84  &54.96  &84.92  &90.91  &62.95  \\
				FGSM attack~\cite{fgsm} & ICLR 2015 & 54.64 & 78.23 & 88.76 & 50.32 & 49.18 & 73.42 & 86.12 & 47.89 \\
				PGD attack~\cite{madry2018towards} & ICLR 2018 & 49.14 & 70.34 & 81.54 & 45.76 & 44.25 & 68.39 & 80.12 & 42.81 \\
				M-FGSM attack~\cite{bai2020adversarial} & TPAMI 2020 &28.48  &64.92  &75.12  &32.88  &25.64  &61.45  &78.31  &30.46  \\
				LTA attack~\cite{gong2022person} & CVPR 2022  &15.16  &56.42  &67.53  &21.47  &19.54  &58.25  &70.72  &24.86  \\
				ODFA attack~\cite{zheng2023u} & IJCV 2023  &26.34  &65.24  &76.92  &30.85  &23.73  &62.46  &73.57  &29.63  \\
				CMPS attack~\cite{gong2024cross} &  NeurIPS 2024 &5.83  &18.14  &27.56  &12.47  &6.42  &19.53  &28.54  &12.23  \\
				MUA attack~\cite{bian2025modality} & TIFS 2025 &4.39  &16.21  &24.87  &5.92  &5.16  &17.05  &26.34  &6.25  \\
				Our attack & NeurIPS 2026   &\textbf{1.97}  &\textbf{9.73}  &\textbf{14.28}  &\textbf{3.01}  &\textbf{3.28}  &\textbf{10.78}  &\textbf{16.12}  &\textbf{4.37}  \\
				\bottomrule
			\end{tabular}
		}
		\label{llcm}
	\end{table*}

	\begin{table}[ht]\scriptsize
		\centering
		\caption{Performance comparison with the state-of-the-art methods in terms of attack transferability in {Visible to Non-visible} and {Non-visible to Visible} scenarios.}
		\label{trans1}
		\begin{tabular}{cc}
			\begin{tabular}{l@{\hskip 8pt}c@{\hskip 8pt}c@{\hskip 8pt}c}
				\hline
				{Cross-Dataset}       & {Method} & {r=1 (\%)} & {mAP (\%)} \\ \hline
				SYSU $\rightarrow$ CnMix  & CMPS  & $13.83 \pm 0.35$ & $12.29 \pm 0.28$ \\
				SYSU $\rightarrow$ CnMix  & MUA         & $11.24 \pm 0.26$ & $10.18 \pm 0.22$ \\
				SYSU $\rightarrow$ CnMix  & Ours                        & $\mathbf{6.42 \pm 0.19}$ & $\mathbf{5.74 \pm 0.14}$ \\
				\midrule
				SYSU $\rightarrow$ Sketch & CMPS  & $16.46 \pm 0.40$ & $14.78 \pm 0.32$ \\
				SYSU $\rightarrow$ Sketch & MUA         & $13.53 \pm 0.29$ & $12.26 \pm 0.24$ \\
				SYSU $\rightarrow$ Sketch & Ours                       & $\mathbf{6.73 \pm 0.21}$ & $\mathbf{6.41 \pm 0.17}$ \\ \hline
			\end{tabular} &
			\begin{tabular}{l@{\hskip 8pt}c@{\hskip 8pt}c@{\hskip 8pt}c}
				\hline
				{Cross-Dataset}       & {Method} & {r=1 (\%)} & {mAP (\%)} \\ \hline
				CnMix $\rightarrow$ SYSU  & CMPS  & $15.75 \pm 0.38$ & $14.41 \pm 0.30$ \\
				CnMix $\rightarrow$ SYSU  & MUA         & $12.02 \pm 0.26$ & $11.15 \pm 0.21$ \\
				CnMix $\rightarrow$ SYSU  & Ours                       & $\mathbf{7.93 \pm 0.20}$ & $\mathbf{7.44 \pm 0.17}$ \\
				\midrule
				Sketch $\rightarrow$ SYSU & CMPS  & $17.34 \pm 0.45$ & $16.87 \pm 0.35$ \\
				Sketch $\rightarrow$ SYSU & MUA        & $14.20 \pm 0.33$ & $13.09 \pm 0.27$ \\
				Sketch $\rightarrow$ SYSU & Ours                       & $\mathbf{9.11 \pm 0.25}$ & $\mathbf{8.39 \pm 0.19}$ \\ \hline
			\end{tabular} \\
			(a) {Visible to Non-visible} & (b) {Non-visible to Visible} \\
		\end{tabular}
	\end{table}

	\subsection{Comparison}
	
	We evaluate the proposed method against representative baselines, including AGW~\cite{ye2021deep} and DEEN~\cite{zhang2023diverse}. 
	Tab.~\ref{regdb} and Tab.~\ref{llcm} report the attack performance on RegDB and LLCM datasets.
	
	For a fair comparison in Tab.~\ref{regdb} and Tab.~\ref{llcm}, we adopt a restricted compatible setting of MA. 
	Specifically, no additional surrogate models, external datasets, or extra modalities beyond the evaluated modality-pair setting are introduced during perturbation optimization. 
	All compared methods are evaluated under the same perturbation budget, query-gallery protocol, training split, and modality-pair setting. 
	The perturbation is generated using only the corresponding source training data and the same available ReID model setting as the compared baselines. 
	All hyperparameters are selected according to the original papers or tuned on the source validation split, without using test-set evaluation results. 
	This restricted protocol prevents MA from gaining an advantage from additional modality access or enlarged surrogate-model ensembles, making the comparison focus on the effectiveness of the proposed perturbation optimization strategy itself.

	\begin{wraptable}{r}{0.48\linewidth}
		\vspace{-8pt}
		\centering
		\scriptsize
		\caption{Ablation studies on the DEEN baseline using the LLCM dataset. {Maha.} refers to Mahalanobis distance, {Grad.} represents Gradient-based optimization, {Evol.} indicates Evolutionary Search, and {Struct.} is the proposed Structure-Aware Filtering module.}
		\setlength{\tabcolsep}{2.5pt}
		\renewcommand{\arraystretch}{0.88}
		\begin{tabular}{ccccc|cc}
			\toprule
			No. & Maha. & Grad. & Evol. & Struct. & mAP & rank-1 \\
			\midrule
			0 & $\times$ & $\times$ & $\times$ & $\times$ & 65.84 & 62.53 \\
			1 & $\times$ & $\checkmark$ & $\times$ & $\times$ & 8.54 & 12.35 \\
			2 & $\checkmark$ & $\checkmark$ & $\times$ & $\times$ & 6.88 & 9.47 \\
			3 & $\checkmark$ & $\checkmark$ & $\checkmark$ & $\times$ & 3.43 & 2.49 \\
			4 & $\checkmark$ & $\checkmark$ & $\checkmark$ & $\checkmark$ & \textbf{3.01} & \textbf{1.97} \\
			\bottomrule
		\end{tabular}
		\label{ablation_llcm}
	\end{wraptable}

	Our method consistently achieves the lowest Rank-1 accuracy and mAP under all settings, significantly outperforming prior works. 
	Gradient-based methods such as FGSM~\cite{fgsm}, PGD~\cite{madry2018towards}, LTA~\cite{gong2022person}, and ODFA~\cite{zheng2023u} are limited in cross-modality scenarios because they mainly optimize local gradient directions and do not explicitly model modality-shared structural semantics. 
	CMPS~\cite{gong2024cross} aggregates gradients across the two modalities but relies on Euclidean distance, which overlooks the anisotropic covariance of each modality's feature space. 
	MUA~\cite{bian2025modality} tailors perturbations per modality via dedicated generators, yet this separation can dilute the shared perturbation signal that facilitates cross-modal transfer. 
	In contrast, MA strengthens attack effectiveness on the fixed pair through a Mahalanobis-guided universal direction that respects feature covariance, followed by sparse evolutionary refinement that corrects biases without being confined to the gradient signals of either modality alone.

	By contrast, even under the restricted compatible setting used in Tab.~\ref{regdb} and Tab.~\ref{llcm}, our method improves attack transferability by combining gradient-based universal perturbation learning with structure-aware evolutionary sparse refinement. 
	The first stage captures a modality-shared attack direction, while the second stage selectively refines sparse perturbations according to attention-shift and retrieval-based fitness signals. 
	This design enables effective perturbation optimization without relying on additional surrogate models, external datasets, or extra modality information in these comparisons.

	\subsection{Ablation Study}
	\textbf{Experiments Comparing Attack Transferability.} 
	We verify transferability using heterogeneous datasets, including SYSU, RegDB, Sketch, and CnMix. In our transfer attack experiments, each dataset is trained with a different model to represent a specific modality. For instance, in Tab.~\ref{trans1}, SYSU$\rightarrow$CnMix indicates that the universal perturbation is optimized on SYSU, with RegDB and Sketch used for auxiliary optimization, and then transferred to CnMix for testing.  
	\begin{wrapfigure}{r}{0.48\linewidth}
		\centering
		\includegraphics[width=0.95\linewidth]{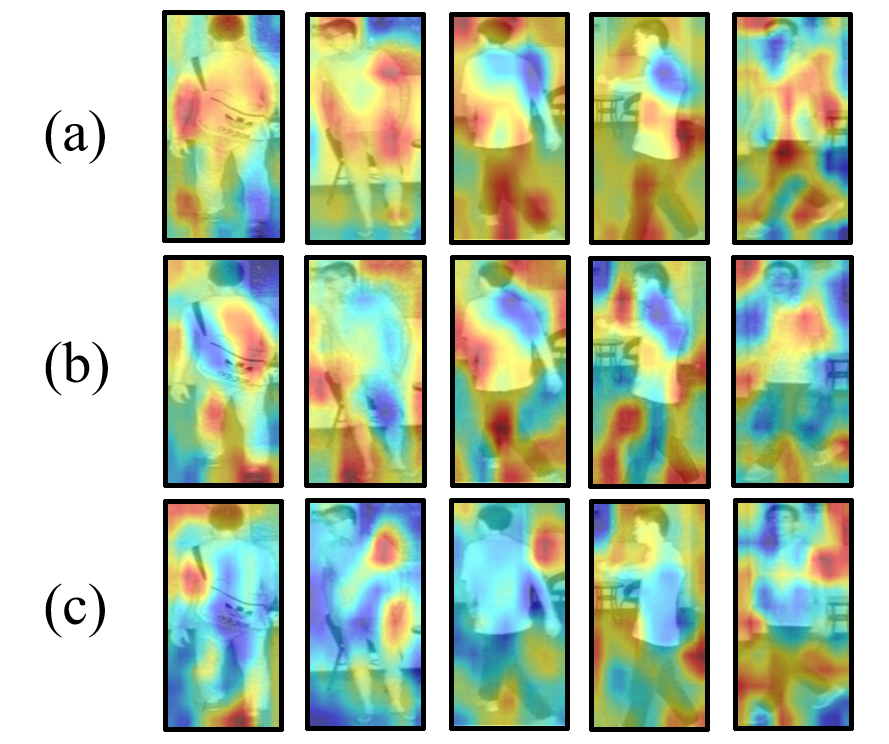}
		\caption{
			Visualization of attention heatmaps under different settings. (a) Clean samples without adversarial perturbation. (b) Perturbed samples generated by MA without structure-aware filtering. (c) Perturbed samples generated by MA with structure-aware filtering.
		}
		\label{heatmap}
	\end{wrapfigure}
	From the results, CMPS and MUA both degrade under heterogeneous cross-dataset transfer. CMPS~\cite{gong2024cross} jointly optimizes perturbations by aggregating gradients across a fixed modality pair, improving cross-modal transfer over single-modality attacks. However, its optimization remains coupled to the source modality pair, potentially leading to source-pair-specific overfitting and limiting its effectiveness on unseen modality combinations. MUA~\cite{bian2025modality} improves scalability by assigning a dedicated generator to each modality, but this design uses modality-specific generators trained on accessible modalities and incurs computational overhead that grows with the number of modalities. In contrast, MA decouples perturbation generation from fixed source pairs: the universal perturbation captures a modality-shared attack direction, while the evolutionary refinement is designed to mitigate source-pair-specific bias through multi-objective search over heterogeneous models. In this heterogeneous transfer setting, all compared methods, including CMPS and MUA, are optimized using the same accessible source and non-target auxiliary domains, ensuring a matched multi-source comparison. Therefore, the performance improvement reflects the advantage of the proposed optimization strategy rather than additional data access. This two-stage design aims to reduce reliance on any single source pair, while Table~\ref{trans1} reports improved transfer to evaluated held-out domains.

	Tab.~\ref{ablation_llcm} summarizes the contribution of each component in our framework. No.0 denotes the clean baseline without perturbation. Introducing only the gradient-based universal perturbation (No.1) already degrades performance, but the gain remains limited. Replacing Euclidean distance with Mahalanobis distance (No.2) further reduces Rank-1 and mAP, confirming its advantage in capturing modality-aware feature structures. Adding evolutionary search (No.3) brings a clear improvement, {showing improved attack effectiveness on this known modality pair}. Finally, incorporating the structure-aware filtering module (No.4) achieves the best attack performance, {showing a further gain in this LLCM ablation}.

	We further visualize attention heatmaps under different configurations in Fig.~\ref{heatmap}. Compared with clean samples and MA without structure-aware filtering, MA with structure-aware filtering {shows larger visualized attention shifts in these examples}. {These examples illustrate attention changes associated with filtering; cross-dataset transfer is evaluated separately in Table~\ref{trans1}}.
	Additional studies on evolutionary hyperparameters, perturbation budgets, modality numbers, cross-architecture transferability, and adversarial defense are provided in the supplementary materials. These results {provide additional observations of MA in the evaluated settings}.

	\section{Conclusion}
	This paper introduced MA, a cross-modal transferable adversarial attack framework for person re-identification. 
	MA is designed to address potential source-pair overfitting by decomposing the attack optimization into a Mahalanobis-guided universal perturbation and a structure-aware sparse correction, optimized through gradient-based learning and evolutionary refinement, respectively.
	This two-stage design {shifts the optimization focus from source-pair effectiveness toward cross-modal transferability}. 
	Extensive experiments across multiple datasets, modalities, and model architectures {show improved attack transferability in the evaluated benchmark, held-out-domain, and cross-architecture settings}.

	\section*{Acknowledgment}
	This work was supported in part by the National Natural Science Foundation of China under Grants 52535009, 62276222, and 62276221, and by the China Scholarship Council under Grant No.~202506310113.

	\bibliographystyle{unsrt} \small
	\bibliography{main}

	%%%%%%%%%%%%%%%%%%%%%%%%%%%%%%%%%%%%%%%%%%%%%%%%%%%%%%%%%%%%
	
	% Define a new command for the large title
	\newcommand{\suptitle}[1]{%
		\begin{center}
			\Large\bfseries #1
		\end{center}
	}
	
	\clearpage
	
	\suptitle{Supplemental Material}
	
	% 补充材料目录
	\begin{minipage}{\linewidth}
		\tableofcontents  % 为补充材料创建目录
	\end{minipage}
	\newpage

	\section{Analysis of Evolutionary Search} \label{A}
	Here, we simplify the problem and appropriately reformulate it to facilitate analysis.
	\subsection{Objective of Sparse Perturbations}
	The goal of introducing sparse perturbations is to minimize the number of changes made to the input while maximizing the error induced across various models. The fitness function, which evaluates the effectiveness of a perturbation across multiple models, is central to this process.
	
	\subsection{Fitness Function}
	Let $\Phi$ represent a collection of models $\{M_1, M_2, \ldots, M_k\}$, where each model $M_i$ has an associated misclassification rate $r_i(x)$ for an input $x$. The fitness function $f(x)$ for a perturbation vector $x$ is defined as:
	\begin{equation}
		f(x) = \sum_{i=1}^k w_i \cdot r_i(x) - \lambda_s \cdot \|x\|_0
	\end{equation}
	Here, $\|x\|_0$ denotes the sparsity of the perturbation vector (the number of non-zero elements), $w_i$ is the weight associated with model $M_i$, reflecting its importance in the overall fitness, and $\lambda_s$ is a regularization parameter that controls the significance of sparsity.
	
	\subsection{Generation of Sparse Perturbations}
	Evolutionary algorithms utilize selection, crossover, and mutation operations to generate new perturbations. The mutation operation, designed to maintain sparsity while enhancing fitness, is defined as:
	\begin{equation}
		x' = x + \delta
	\end{equation}
	where $\delta$ is a small change vector, typically non-zero in only a few components of $x$. $\delta$ is chosen to maximize $f(x + \delta)$.
	
	\subsection{Convergence Analysis}
	By appropriately choosing the parameters $\lambda_s$ and $w_i$, evolutionary search can converge to high-quality sparse perturbations. Theoretically, consider the change in the fitness function over iterations, assuming:
	\begin{equation}
		\lim_{t \to \infty} f(x^{(t)}) = f^*
	\end{equation}
	where $x^{(t)}$ is the perturbation vector after the $t$-th iteration, and $f^*$ represents the optimal achievable fitness.
	
	To ensure that each iteration does not decrease the fitness, we need:
	\begin{equation}
		f(x^{(t+1)}) \geq f(x^{(t)})
	\end{equation}
	This indicates that the fitness is non-decreasing over iterations, suggesting that the algorithm is at least locally optimal. By designing \(\delta\) to ensure that \(f(x + \delta) > f(x)\), this can be achieved. This can be specifically implemented through techniques such as non-dominated sorting to select the most effective perturbations. Please note that our goal is to fine-tune the universal perturbation using the search process; in fact, we do not need to find the perturbation with the optimal fitness value.
	
	Through the analysis provided, we theoretically demonstrate that evolutionary search can generate effective sparse perturbations that are effective across multiple models.

	\section{Theoretical Motivation for Structure-Aware Filtering}
	
	In this section, we provide a formal motivation for the attention-shift-based filtering module introduced in Sec.~3. Attention shift serves as a structure-aware proxy metric for preliminarily filtering sparse perturbation candidates before multi-objective fitness evaluation. The final selection of perturbations is still determined by the attack-oriented fitness, while attention shift is used to discard candidates that are unlikely to affect the model's discriminative feature extraction process.
	
	\subsection{Attention Shift as a Proxy for Structural Sensitivity}
	
	Let $f_\theta(x) \in \mathbb{R}^d$ denote the feature embedding produced by a ReID model $M_\theta$ for an input image $x$. Many ReID models extract identity-discriminative features from spatially distributed person regions. Let $A_\theta(x) \in \mathbb{R}_{+}^{h \times w}$ denote a normalized attention or saliency map that indicates the spatial regions contributing to the embedding. Such a map can be obtained from an explicit attention module, transformer attention, or gradient-based visualization methods such as Grad-CAM.
	
	We interpret $A_\theta(x)$ as a proxy for spatial feature sensitivity rather than an exact measure of the input-output Jacobian. Intuitively, a sparse perturbation that does not change the attended regions is less likely to modify the identity-related feature aggregation process, whereas a perturbation that induces a clear attention deviation is more likely to affect the regions used for ReID matching.
	
	To make this intuition concrete, consider a simplified attention-weighted representation:
	\[
	f_\theta(x) = \sum_{p=1}^{N} A_\theta(x)[p] h_\theta(x)[p],
	\]
	where $p$ indexes spatial locations, $A_\theta(x)[p]$ is the normalized attention weight at location $p$, and $h_\theta(x)[p]$ is the corresponding local descriptor. For a sparse perturbation $\eta$, the embedding variation can be decomposed as
	\[
	f_\theta(x+\eta)-f_\theta(x)
	=
	\sum_{p=1}^{N}
	\Delta A_\theta[p] h_\theta(x)[p]
	+
	\sum_{p=1}^{N}
	A_\theta(x+\eta)[p] \Delta h_\theta[p],
	\]
	where
	\[
	\Delta A_\theta[p] = A_\theta(x+\eta)[p]-A_\theta(x)[p],
	\]
	and
	\[
	\Delta h_\theta[p] = h_\theta(x+\eta)[p]-h_\theta(x)[p].
	\]
	This decomposition shows that attention deviation can directly contribute to embedding variation through the first term, while local descriptor changes contribute through the second term. Therefore, attention shift provides a useful proxy for identifying sparse perturbations that interact with the model's identity-related feature aggregation process. However, attention shift alone is not sufficient to guarantee retrieval failure, which is why MA uses it only as a filtering signal before attack-fitness evaluation.
	
	\subsection{Cross-Model Attention Alignment Assumption}
	
	For attention-shift filtering to be useful for transfer attacks, we assume that different ReID models trained on related cross-modal tasks rely partially on overlapping person-related regions. This assumption does not require identical attention maps across models or modalities. It only requires a weak positive alignment in expectation.
	
	\textbf{Assumption 1. Cross-model structural alignment.}
	Let $A_s(x)$ and $A_t(x)$ denote normalized attention maps of a source model $M_s$ and an unseen target model $M_t$, respectively. There exists a task-dependent constant $\rho>0$ such that
	\[
	\mathbb{E}_{x}
	\left[
	\operatorname{sim}(A_s(x), A_t(x))
	\right]
	\ge \rho,
	\]
	where $\operatorname{sim}(\cdot,\cdot)$ denotes a similarity measure such as cosine similarity.
	
	This assumption is plausible in cross-modal ReID because different models ultimately need to extract person-related cues, such as body silhouette, part layout, and discriminative local regions, even when the image modalities differ. Under this assumption, a sparse perturbation that consistently changes source-model attention is more likely to affect regions that are also relevant to unseen target models.
	
	\subsection{Attention-Shift Filtering as a Surrogate Selection Objective}
	
	Consider a population of sparse perturbation candidates $\mathcal{P}=\{\eta_i\}$. For each candidate $\eta_i$, MA computes an attention-shift score over the source ensemble:
	\[
	S(\eta_i)
	=
	\mathbb{E}_{M_s \in \mathcal{M}_s}
	\left[
	\|A_s(x+\delta+\eta_i)-A_s(x)\|_1
	\right],
	\]
	where $\delta$ is the universal perturbation learned in the first stage. The filtering module retains candidates with large $S(\eta_i)$ before applying the final multi-objective fitness evaluation.
	
	Under Assumption 1, a larger source attention-shift score indicates that the candidate perturbation is more likely to affect person-related regions shared across different ReID models. Thus, $S(\eta_i)$ can be viewed as a surrogate indicator of structural transferability. Importantly, this does not imply that maximizing $S(\eta_i)$ alone maximizes the final attack success rate. Attention shift only biases the search toward candidates that are structurally meaningful, while the final evolutionary selection still depends on the attack-oriented fitness, including retrieval degradation, sparsity, and structural impact.
	
	Therefore, attention-shift filtering should be understood as a principled proxy for improving the efficiency and reliability of sparse evolutionary refinement, rather than as a deterministic theoretical guarantee. It helps remove candidates that perturb insensitive regions and concentrates the search on sparse corrections that are more likely to influence identity-related feature extraction across heterogeneous models.

	\section{Stability Analysis of Evolutionary Refinement}
	
	We analyze the stability property of the evolutionary refinement stage on the set of substitute models. Let $\delta_u$ denote the universal perturbation from the first stage and $\eta$ denote the sparse refinement. For theoretical tractability, we consider a scalarized approximation of the actual multi-objective objective (the actual implementation uses Pareto-based non-dominated sorting with the multi-objective vector in Eq.~(9); the scalarized form below is used only for analysis):
	
	\begin{equation}
		f(\eta)=\sum_{i=1}^{n}w_i r_i(\delta_u+\eta)-\lambda_s\|\eta\|_0, \tag{G.1}
	\end{equation}
	
	where $n$ is the number of substitute models, $r_i(\cdot)$ measures attack effectiveness on model $M_i$, $w_i\geq0$ is the model-specific weight, and $\lambda_s>0$ controls the sparsity penalty.
	
	Define the weighted cross-model attack effectiveness as:
	
	\begin{equation}
		R(\eta)=\sum_{i=1}^{n}w_i r_i(\delta_u+\eta). \tag{G.2}
	\end{equation}
	
	To quantify the complementary contribution of the sparse refinement over the universal perturbation, define the aggregate complementarity measure as:
	
	\begin{equation}
		\bar{\alpha}(\eta)=\frac{R(\eta)-R(\mathbf{0})}{R(\mathbf{0})+\tau}, \tag{G.3}
	\end{equation}
	
	where $R(\mathbf{0})=\sum_i w_i r_i(\delta_u)$ is the attack effectiveness of the universal perturbation alone, and $\tau>0$ is a small constant to avoid division by zero.
	
	Let $\mathcal{A}^{(g)}$ denote the elitist archive after the $g$-th evolutionary generation, which stores the best perturbation candidates discovered so far. Under elitist selection, previous high-quality candidates are preserved, thus:
	
	\begin{equation}
		\mathcal{A}^{(g)}\subseteq \mathcal{A}^{(g+1)}. \tag{G.4}
	\end{equation}
	
	Define the best-so-far sparse refinement in the archive with respect to the weighted cross-model attack effectiveness as:
	
	\begin{equation}
		\eta_*^{(g)} = \arg\max_{\eta\in\mathcal{A}^{(g)}} R(\eta). \tag{G.5}
	\end{equation}
	
	Since the archive at generation $g+1$ contains all candidates from generation $g$, we have:
	
	\begin{equation}
		R(\eta_*^{(g+1)}) = \max_{\eta\in\mathcal{A}^{(g+1)}} R(\eta) \;\geq\; \max_{\eta\in\mathcal{A}^{(g)}} R(\eta) = R(\eta_*^{(g)}). \tag{G.6}
	\end{equation}
	
	Because $R(\mathbf{0})$ and $\tau$ are fixed during the evolutionary refinement stage, substituting (G.6) into (G.3) yields:
	
	\begin{equation}
		\bar{\alpha}^{(g+1)} \geq \bar{\alpha}^{(g)}. \tag{G.7}
	\end{equation}
	
	Thus, the best-so-far attack effectiveness on the substitute model set is non-decreasing over generations, and the aggregate complementarity is also non-decreasing. This stability property guarantees that the refinement process does not degrade the already learned universal perturbation on the models used during optimization.
	
	Because the feasible sparse perturbation space is finite under fixed $\ell_0$ and $\ell_\infty$ constraints, and because $r_i(\cdot)$ is bounded, the sequence $R(\eta_*^{(g)})$ converges to a stable value. This should be understood as a practical stability property of elitist evolutionary refinement.

	\section{Detailed Implementation and Hyperparameters}
	\label{app:hyperparams}
	
	This section provides implementation details for the proposed two-stage MA framework.
	Table~\ref{tab:hyper_ma} lists the hyperparameters used in our experiments.
	Unless otherwise specified, these values are used across all datasets and settings.
	In the first stage, the universal perturbation $\delta$ is optimized by 
	Mahalanobis-guided gradient updates. In the second stage, $\delta$ is fixed, 
	and the sparse correction $\eta$ is optimized by structure-aware evolutionary refinement.
	
	\begin{table}[htbp]
		\centering
		\caption{Hyperparameters of the proposed MA framework.}
		\label{tab:hyper_ma}
		\begin{tabular}{lll}
			\hline
			Parameter & Value & Description \\
			\hline
			$\epsilon$ & 8 & $\ell_\infty$ perturbation budget (pixel scale 0--255) \\
			$\alpha$ & $\epsilon/10 = 0.8$ & Learning rate for gradient update \\
			$\beta$ & 0.9 & Momentum decay coefficient \\
			$T$ & 30 & Number of gradient iterations in Stage 1 \\
			$\rho$ & 0.3 & Margin in the triplet loss (Eq.~3) \\
			$\lambda$ & $10^{-4}$ & Diagonal regularization for the covariance matrix (Eq.~5) \\
			$k$ & 20 & Maximum number of perturbed pixel locations in $\eta$ \\
			$P$ & 4 & Population size in evolutionary search \\
			$G$ & 200 & Number of evolutionary generations in Stage 2 \\
			$p_c$ & 0.3 & Crossover probability \\
			$p_m$ & 0.2 & Mutation probability \\
			\hline
		\end{tabular}
	\end{table}
	
	\textbf{Choice of $\rho$ and $\lambda$.}
	The margin $\rho=0.3$ is selected via grid search on the validation split of RegDB 
	(visible-thermal) to balance attack effectiveness and feature-space regularization.
	The covariance regularization coefficient $\lambda=10^{-4}$ ensures numerical stability 
	when inverting the covariance matrix without dominating the diagonal term.
	
	\textbf{Evolutionary refinement budget.}
	In the revised two-stage implementation, evolutionary refinement is performed only after 
	the universal perturbation $\delta$ has been learned and fixed. Therefore, $G$ denotes the 
	total number of evolutionary generations in Stage 2.

	\section{Multiform Attack Optimization Framework}
	
	\begin{algorithm}[t]
		\small
		\caption{Multiform Attack Optimization Framework}
		\label{algorithm}
		\begin{algorithmic}[1]
			\State \textbf{Input:} 
			$\mathcal{D}_{\text{train}}$, $\mathcal{D}_{\text{evo}}$,
			$\mathcal{M}_{\text{grad}}$, $\mathcal{M}_{\text{evo}}$,
			perturbation budget $\epsilon$,
			gradient steps $T$,
			population size $P$,
			generations $G$,
			sparsity budget $k$,
			attention-filter size $K$.
			\State \textbf{Output:} adversarial perturbation $\Delta$.
			
			\Statex
			\State \textbf{Stage 1: Universal Perturbation Learning}
			\State Initialize $\delta \leftarrow \mathbf{0}$ and momentum $v \leftarrow \mathbf{0}$.
			\For{$t=1$ \textbf{to} $T$}
			\State Sample mini-batch $x \sim \mathcal{D}_{\text{train}}$.
			\State Compute Mahalanobis-guided loss 
			$\mathcal{L}_{\text{tri}}(\delta,x)$ using $\mathcal{M}_{\text{grad}}$.
			\State Update $\delta$ by momentum-based gradient optimization and project it onto 
			$\|\delta\|_{\infty}\leq\epsilon$.
			\EndFor
			
			\Statex
			\State \textbf{Stage 2: Sparse Evolutionary Refinement}
			\State Fix $\delta$ and initialize a population 
			$\mathcal{S}^{(0)}=\{\eta_i\}_{i=1}^{P}$ of $k$-sparse residuals.
			\State $\mathcal{S}^{(0)}\leftarrow
			\textsc{Repair}(\mathcal{S}^{(0)},k,\delta,\epsilon)$.
			\State Initialize archive $\mathcal{A}\leftarrow\emptyset$ and 
			$\eta_{\text{best}}\leftarrow\mathbf{0}$.
			
			\For{$g=1$ \textbf{to} $G$}
			\State Sample mini-batch $x_{\text{evo}}\sim\mathcal{D}_{\text{evo}}$.
			\State Select top-$K$ parents 
			$\mathcal{P}^{(g)}$ from $\mathcal{S}^{(g-1)}$ by attention-shift scores.
			\State Generate offspring $\mathcal{O}^{(g)}$ from $\mathcal{P}^{(g)}$ 
			via crossover and mutation.
			\State Repair all offspring to satisfy 
			$\|\delta+\eta\|_{\infty}\leq\epsilon$ and $\|\eta\|_0\leq k$.
			\State Form candidate set 
			$\mathcal{C}^{(g)}
			=\mathcal{P}^{(g)}\cup\mathcal{O}^{(g)}\cup\{\eta_{\text{best}}\}$.
			\State Evaluate multi-objective fitness $\Phi(\eta)$ for each 
			$\eta\in\mathcal{C}^{(g)}$ on $\mathcal{M}_{\text{evo}}$.
			\State Update archive 
			$\mathcal{A}\leftarrow\textsc{UpdateArchive}(\mathcal{A}\cup\mathcal{C}^{(g)})$.
			\State Update best sparse correction
			$\eta_{\text{best}}\leftarrow\textsc{SelectPriority}(\mathcal{A})$.
			\State Set 
			$\mathcal{S}^{(g)}
			\leftarrow
			\textsc{Truncate}(
			\textsc{NonDominatedSort}(\mathcal{C}^{(g)}),P)$.
			\EndFor
			
			\State $\Delta \leftarrow \Pi_{\|\cdot\|_{\infty}\leq\epsilon}
			(\delta+\eta_{\text{best}})$.
			\State \Return $\Delta$.
		\end{algorithmic}
	\end{algorithm}

	To construct perturbations with strong cross-modal transferability and high discriminative impact, 
	we introduce a two-stage multiform attack optimization framework consisting of 
	(i) gradient-driven universal perturbation optimization and 
	(ii) multi-generation evolutionary sparse refinement. 
	The complete procedure is summarized in Algorithm~\ref{algorithm}.
	
	\paragraph{Stage 1: Gradient-Based Universal Perturbation Optimization.}
	At each iteration, a mini-batch is sampled from the source training set 
	$\mathcal{D}_{\text{train}}$, and the universal perturbation $\delta$ is updated using 
	the gradient model $\mathcal{M}_{\text{grad}}$.
	Specifically, we compute the Mahalanobis-guided triplet loss 
	$\mathcal{L}_{\text{tri}}(\delta, x)$ and optimize $\delta$ through a momentum-based 
	update followed by a projection onto the perturbation bound $[-\epsilon, \epsilon]$. 
	This stage produces a universal feature-disruptive perturbation that captures a coarse 
	cross-modal attack direction, providing a strong initialization for the subsequent 
	structural refinement.

	For each source modality $m$, we precompute class centroids from the training set. Let $\mathcal{X}_c^m$ denote the set of training samples of identity $c$ in modality $m$. The centroid for identity $c$ is computed as:
	\[
	\mu_c^m = \frac{1}{|\mathcal{X}_c^m|} \sum_{x \in \mathcal{X}_c^m} F(x),
	\]
	where $F$ is the feature extractor.
	
	Given a query image $x_i$ with ground-truth identity $y_i$, the positive centroid $C_p^m(x_i)$ is the centroid of identity $y_i$. The negative centroid $C_n^m(x_i)$ is selected as the centroid with the smallest Mahalanobis distance to $f_i$ among all identities different from $y_i$ (hard negative mining). This strategy is applied for both source modalities $m_1$ and $m_2$.

	\paragraph{Stage 2: Multi-Generation Evolutionary Sparse Refinement.}
	To further enhance the structural relevance and model-transferability of the perturbation, 
	after completing the gradient-based universal perturbation learning stage, 
	we fix the learned universal perturbation $\delta$ and then perform evolutionary sparse refinement using mini-batches sampled from $\mathcal{D}_{\text{evo}}$.
	
	The construction of $D_{\text{train}}$ and $D_{\text{evo}}$ depends on the evaluation protocol. 
	In the restricted compatible setting, no auxiliary domain is used, and both $D_{\text{train}}$ and $D_{\text{evo}}$ are sampled from the training split of the evaluated source modality pair. 
	In the heterogeneous transfer setting, $D_{\text{train}}$ is sampled from the primary source training split, while $D_{\text{evo}}$ is sampled from the union of the primary source training split and all accessible non-target auxiliary training splits. 
	We use balanced sampling across domains when constructing mini-batches from $D_{\text{evo}}$ so that no auxiliary domain dominates the evolutionary fitness evaluation. 
	The held-out target domain is never included in either $D_{\text{train}}$ or $D_{\text{evo}}$.

	\textbf{Population Initialization.}
	We initialize a population $\mathcal{S}^{(0)} = \{\eta_i\}_{i=1}^{P}$, where each 
	$\eta_i$ is a $k$-sparse corrections perturbation with entries in $\{-1,0,+1\}$.
	Specifically, each individual contains at most $k$ perturbed pixel locations; the 
	channel-wise correction values at selected locations are sampled from $\{-1,0,+1\}$, 
	while all unselected locations are set to zero. 
	Such discrete sparse candidates promote structured refinement in localized, 
	discriminative regions while keeping the correction compact.
	
	\textbf{Attention-Shift-Based Structural Filtering.}
	For each generation $g = 1 \dots G$, the structural impact of each individual $\eta_i$ 
	is evaluated through the attention shift induced on models in $\mathcal{M}_{\text{evo}}$:
	\begin{equation}
		s_i = \|A_{\text{adv}} - A_{\text{clean}}\|_1.
	\end{equation}
	Candidates are ranked according to $s_i$, and only the top-$K$ individuals are retained. 
	This mechanism serves as a structure-aware filtering strategy, ensuring that the 
	evolutionary process focuses on perturbations that meaningfully disrupt discriminative 
	attention regions.
	
	\textbf{Genetic Variation and Multi-Objective Selection.} 
	The selected individuals undergo crossover and mutation to generate a new set of offspring.  
	After genetic variation, a repair operation is applied to ensure feasibility: it keeps at 
	most $k$ perturbed pixel locations, sets all other entries to zero, and guarantees that 
	the correction values remain in $\{-1,0,+1\}$.  
	If a correction value would violate the global perturbation constraint, i.e.,  
	$|\delta[p,c]+\eta[p,c]|>\epsilon$, that correction value is removed.  
	Therefore, each candidate satisfies $\|\eta\|_0 \le k$ and  
	$\|\delta+\eta\|_\infty \le \epsilon$ before fitness evaluation.  
	A non-dominated sorting procedure is then applied to select the next-generation population  
	$\mathcal{S}^{(g)}$, preserving a balance between attack effectiveness and perturbation  
	compactness.
	
	\paragraph{Final Perturbation Assembly.}
	After $G$ generations, the best-ranked individual $\eta_{\text{best}}$ is selected and 
	added to the universal perturbation, yielding a refined multiform adversarial perturbation:
	\begin{equation}
		\Delta = \mathrm{clip}(\delta + \eta_{\text{best}}, -\epsilon, \epsilon).
	\end{equation}
	Through the combination of gradient-guided universal optimization and structure-aware 
	evolutionary sparse refinement, our framework produces perturbations that are universal, 
	sparse, structurally discriminative, and transferable across models and modalities. $\eta_{\mathrm{best}}$ is a sparse perturbations shared across all queries, rather than a query-specific correction. Once selected from the evolutionary archive using source mini-batches and surrogate models, it is fixed and combined with $\delta$ to form the final universal perturbation $\Delta=\Pi_{\epsilon}(\delta+\eta_{\mathrm{best}})$, which is directly applied to unseen queries without per-query re-optimization or target-model queries.

	\section{Details of the Evolutionary Algorithm}  \label{Crossover}
	\subsection{Initialization}
	The attack method initializes by setting the number of perturbed pixels to $k$ and constructing a set $\mathcal{P}$ of $s$ solutions. Each solution is created by uniformly sampling pixel locations from the set $\{1, \cdots, h \cdot w\}$. The initial perturbation values for each color channel are randomly sampled from the set $\{-1, 0, 1\}$:
	
	\begin{equation}
		\mathcal{P} = \{\mathcal{P}_1, \mathcal{P}_2, \ldots, \mathcal{P}_s\}, \quad \mathcal{P}_i \in \{-1, 0, 1\}
	\end{equation}
	
	where:
	\begin{itemize}
		\item $\mathcal{P}$: The set of initial solutions.
		\item $s$: The total number of solutions.
	\end{itemize}
	When invoking the evolutionary process, we set a fixed number of iterations to 200.

	\subsection{Crossover}
	The crossover operation aims to generate new solutions by combining traits from two parent solutions. This process helps in exploring the solution space and inheriting beneficial traits from the parents.
	
	Crossover combines traits from two parent solutions $P_a$ and $P_b$ to generate new solutions:
	
	\begin{equation}
		M' = (M_a \cup M_b) \setminus (M_a \cap M_b)
	\end{equation}
	\begin{equation}
		\Delta' = \Delta_a \cup \Delta_b
	\end{equation}
	
	where:
	\begin{itemize}
		\item $M_a$, $M_b$: Sets of pixel locations from the parent solutions $P_a$ and $P_b$, respectively.
		\item $M'$: The set of pixel locations for the new solution.
		\item $\Delta_a$, $\Delta_b$: Sets of perturbation values corresponding to $M_a$ and $M_b$.
		\item $\Delta'$: The set of perturbation values for the new solution.
	\end{itemize}
	
	\subsection{Mutation}
	The mutation operation introduces variations by modifying a subset of pixel locations. This process helps in exploring new regions of the solution space and maintaining diversity among solutions.
	
	Mutation introduces variations to solutions by modifying a subset of pixel locations:
	
	\begin{equation}
		M'' = (M' \setminus A) \cup B, \quad \Delta'' = (\Delta' \setminus \Delta_A) \cup \Delta_B
	\end{equation}
	
	where:
	\begin{itemize}
		\item $M'$: The set of pixel locations before mutation.
		\item $A$: A randomly selected subset of pixel locations to be replaced.
		\item $B$: A new subset of pixel locations to be introduced (obtained through random selection).
		\item $M''$: The set of pixel locations after mutation.
		\item $\Delta'$: The set of perturbation values before mutation.
		\item $\Delta_A$: The perturbation values corresponding to $A$.
		\item $\Delta_B$: The perturbation values corresponding to $B$.
		\item $\Delta''$: The set of perturbation values after mutation.
	\end{itemize}

	\subsection{Evaluation}
	
	The evaluation process calculates the objective vector 
	$\Phi(f_{\mathrm{adv}})=(\tilde{\mathcal{D}}(f_{\mathrm{adv}}), 
	\tilde{\mathcal{S}}(f_{\mathrm{adv}}), \|\eta\|_2)^\top$ 
	for each candidate solution to measure its quality. 
	Here, $\tilde{\mathcal{D}}(f_{\mathrm{adv}})$ measures the transformed metric objective, 
	$\tilde{\mathcal{S}}(f_{\mathrm{adv}})$ measures the transformed retrieval attack success objective, 
	and $\|\eta\|_2$ encourages compact sparse refinements. 
	The retrieval attack success is computed under the standard ReID inference protocol, where the adversarial query is used to retrieve gallery images and success is determined by whether the top-ranked gallery identity differs from the ground-truth query identity. 
	The $\ell_0$ sparsity of each perturbation is controlled by the constraint $\|\eta\|_0 \leq k$.

	\subsection{Selection}
	
	The selection stage aims to preserve sparse perturbation candidates that achieve strong cross-model retrieval transferability while maintaining perturbation compactness. 
	Following the main paper, each candidate perturbation $\eta$ is evaluated by the objective vector
	\begin{equation}
		\Phi(\eta)
		=
		\left(
		\tilde{\mathcal{D}}(f_{\mathrm{adv}}),
		\tilde{\mathcal{S}}(f_{\mathrm{adv}}),
		\|\eta\|_2
		\right)^\top ,
	\end{equation}
	where all objectives are minimized. 
	Here, $\tilde{\mathcal{D}}(f_{\mathrm{adv}})$ denotes the transformed metric objective, 
	$\tilde{\mathcal{S}}(f_{\mathrm{adv}})=1-\mathcal{S}(f_{\mathrm{adv}})$ denotes the transformed retrieval attack success objective, 
	and $\|\eta\|_2$ encourages compact sparse refinements. 
	Since $\mathcal{S}(f_{\mathrm{adv}})$ is the average retrieval attack success rate over multiple surrogate models, minimizing $\tilde{\mathcal{S}}(f_{\mathrm{adv}})$ is equivalent to maximizing cross-model retrieval transferability.
	
	\paragraph{Pareto dominance.}
	Given two candidate perturbations $\eta_i$ and $\eta_j$ with objective vectors $\Phi_i$ and $\Phi_j$, $\eta_i$ is considered to dominate $\eta_j$ under minimization if
	\begin{equation}
		\Phi_m(\eta_i) \leq \Phi_m(\eta_j), \quad \forall m,
		\quad \text{and} \quad
		\exists m,\ \Phi_m(\eta_i) < \Phi_m(\eta_j).
	\end{equation}
	That is, $\eta_i$ is no worse than $\eta_j$ on all objectives and is strictly better on at least one objective. 
	We apply non-dominated sorting to the union of parent and offspring populations:
	\begin{equation}
		\mathcal{P}'=\mathrm{NonDominatedSort}(\mathcal{P}\cup\mathcal{O}),
	\end{equation}
	where $\mathcal{P}$ and $\mathcal{O}$ denote the parent and offspring populations, respectively. 
	This elitist selection preserves high-quality candidates across generations.
	
	\paragraph{Transferability-oriented priority.}
	The Pareto domination rule provides the general multi-objective selection criterion. 
	In implementation, when selecting the final sparse refinement from the best-ranked front, we further use a transferability-oriented priority rule:
	\begin{enumerate}
		\item If candidate $\eta_i$ achieves a higher average retrieval attack success rate than candidate $\eta_j$, i.e.,
		$\mathcal{S}(f_{\mathrm{adv}}^{\,i}) > \mathcal{S}(f_{\mathrm{adv}}^{\,j})$, then $\eta_i$ is considered more transferable and is preferred.
		
		\item If $\eta_i$ and $\eta_j$ achieve the same average retrieval attack success rate, the candidate with the smaller perturbation magnitude is preferred, i.e.,
		$\|\eta_i\|_2 \leq \|\eta_j\|_2$.
		
		\item If both candidates fail to achieve successful retrieval attacks, i.e., their retrieval attack success rates are both zero, the transformed metric objective is used as a fallback ranking signal, and the candidate with the smaller $\tilde{\mathcal{D}}(f_{\mathrm{adv}})$ is preferred.
	\end{enumerate}
	Here, $f_{\mathrm{adv}}^{\,i}$ and $f_{\mathrm{adv}}^{\,j}$ denote the adversarial feature representations produced by candidates $\eta_i$ and $\eta_j$, respectively.
	This priority rule does not replace Pareto-based non-dominated sorting. 
	Instead, it is used to choose the final sparse refinement from the selected non-dominated candidates. 
	It reflects our practical objective: first maximize cross-model retrieval transferability, then encourage compact sparse perturbations, and finally rely on the metric objective when no candidate has yet achieved successful retrieval transfer.
	
	After non-dominated sorting, the population is ranked into multiple Pareto fronts. 
	The next-generation population is formed by selecting candidates from the best-ranked fronts until the population budget is reached. 
	The final sparse refinement is selected from the best-ranked front according to the priority rule above:
	\begin{equation}
		\eta_{\mathrm{best}}
		=
		\operatorname{Select}_{\mathrm{priority}}
		\left(
		\mathcal{P}_{\mathrm{front}}
		\right),
	\end{equation}
	where $\mathcal{P}_{\mathrm{front}}$ denotes the best-ranked non-dominated front. 
	Specifically, candidates are first ranked by the average retrieval attack success rate $\mathcal{S}(f_{\mathrm{adv}})$, then by the perturbation magnitude $\|\eta\|_2$, and finally by the transformed metric objective $\tilde{\mathcal{D}}(f_{\mathrm{adv}})$ when no candidate achieves successful retrieval transfer.

	\begin{algorithm}[H]\scriptsize
		\caption{Non-Dominating Sorting for Multiform Attack}
		\label{alg:non_dominating_sorting}
		\begin{algorithmic}[1]
			\State \textbf{Input:} Combined population $\mathcal{P}$, objective vectors $\Phi$
			\State \textbf{Output:} Set of fronts $\mathcal{S}_f$
			
			\State $\mathcal{S}_f \gets \{\}$ // Initialize the set of fronts
			\For{$p \in \mathcal{P}$} 
			\State $S_p \gets \{\}$ // Set of $p$ dominated solutions
			\State $n_p \gets 0$ // Domination counter of $p$
			\For{$q \in \mathcal{P}$}
			\If{\Call{Dominates}{$p, q, \Phi$}}  //if p dominates q
			\State $S_p \gets S_p \cup \{q\}$
			\ElsIf{\Call{Dominates}{$q, p, \Phi$}}
			\State $n_p \gets n_p + 1$
			\EndIf
			\EndFor
			\If{$n_p == 0$}
			\State $p_{\text{rank}} \gets 1$ // $p$ belongs to the first front
			\State $\mathcal{S}_{f1} \gets \mathcal{S}_{f1} \cup \{p\}$
			\EndIf
			\EndFor
			\State $i \gets 1$ // Initialize front counter
			\While{$\mathcal{S}_{fi} \neq \emptyset$}
			\State $Q \gets \{\}$ // Store solutions of the next front
			\For{$p \in \mathcal{S}_{fi}$}
			\For{$q \in S_p$}
			\State $n_q \gets n_q - 1$
			\If{$n_q == 0$}
			\State $q_{\text{rank}} \gets i + 1$
			\State $Q \gets Q \cup \{q\}$
			\EndIf
			\EndFor
			\EndFor
			\State $i \gets i + 1$
			\State $\mathcal{S}_{f(i+1)} \gets Q$
			\EndWhile
			\State \Return $\mathcal{S}_f$
		\end{algorithmic}
	\end{algorithm}
	
	\begin{algorithm}\scriptsize
		\caption{Function to Determine if One Solution Dominates Another}
		\begin{algorithmic}[1]
			
			\Function{Dominates}{$x, y, \Phi$}
			\State $isBetter \gets \text{False}$
			\For{$i \gets 1 \text{ to } \text{length}(\Phi)$}
			\If{$\Phi_i(x) < \Phi_i(y)$}
			\State $isBetter \gets \text{True}$
			\ElsIf{$\Phi_i(x) > \Phi_i(y)$}
			\State \Return \text{False}
			\EndIf
			\EndFor
			\State \Return $isBetter$
			\EndFunction
			
		\end{algorithmic}
	\end{algorithm}

	\begin{algorithm}[t]\small
		\caption{Random Channel Mixing of Grayscale and Sketch Images}
		\begin{algorithmic}[1]
			\State \textbf{Input:} $img$ (Input RGB image), $G$ (Probability of converting to grayscale), $G_{rgb}$ (Probability of mixing grayscale and RGB), $S_{rgb}$ (Probability of mixing sketch and RGB)
			\State \textbf{Output:} $output\_img$ (Transformed image)
			
			\Function{toSketch}{$img$}
			\State $img\_np \gets$ Convert $img$ to numpy array
			\State $img\_inv \gets 255 - img\_np$ \Comment{Invert image colors}
			\State $img\_blur \gets$ Gaussian blur of $img\_inv$
			\State $img\_blend \gets \frac{img\_np}{255 - img\_blur} \times 256$ \Comment{Blend original and blurred images}
			\State \Return Convert $img\_blend$ to image
			\EndFunction
			
			\Function{random\_choose}{$r, g, b, gray\_or\_sketch$}
			\State $p \gets [r, g, b, gray\_or\_sketch, gray\_or\_sketch]$
			\State $idx \gets [0, 1, 2, 3, 4]$
			\State Shuffle $idx$
			\State \Return Merge channels $p[idx[0]], p[idx[1]], p[idx[2]]$ into RGB image
			\EndFunction
			
			\Function{Fuse\_RGB\_Gray\_Sketch}{$img, G, G_{rgb}, S_{rgb}$}
			\State $r, g, b \gets$ Split $img$ into RGB channels
			\State $gray \gets$ Convert $img$ to grayscale
			\State $p \gets$ Random value between 0 and 1
			
			\If{$p < G$}
			\State \Return Merge $gray, gray, gray$ into RGB image
			\ElsIf{$p < G + G_{rgb}$}
			\State $output\_img \gets$ \Call{random\_choose}{$r, g, b, gray$}
			\State \Return $output\_img$
			\ElsIf{$p < G + G_{rgb} + S_{rgb}$}
			\State $sketch \gets$ \Call{toSketch}{$gray$}
			\State $output\_img \gets$ \Call{random\_choose}{$r, g, b, sketch$}
			\State \Return $output\_img$
			\Else
			\State \Return $img$
			\EndIf
			\EndFunction
			
		\end{algorithmic}
		\label{CnMix}
	\end{algorithm}

	\section{Details of the CnMix Processing Algorithm}
\begin{wrapfigure}{r}{0.35\linewidth}  % r=右侧, l=左侧; 宽度设为图片实际宽度的一半
		\vspace{-15pt}  % 可选：减少顶部空白
		\centering
		\includegraphics[width=\linewidth]{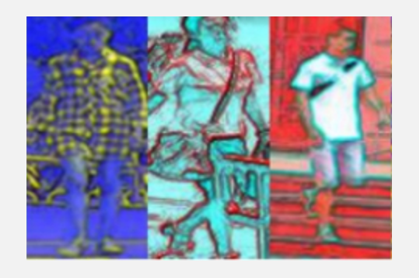}
		\caption{Examples of samples generated by CnMix.}
		\label{picture}
	\end{wrapfigure}

	The final images produced by the CnMix algorithm are shown in Fig.~\ref{picture}. Here we provide the detailed processing workflow of CnMix, which generates new images by randomly mixing the three channels of the images. For specific details, refer to Alg.~\ref{CnMix}.
	
	\paragraph{Sketch Generation.}
	The auxiliary function \texttt{toSketch} converts the input RGB image into a sketch-like version using a classic color-inversion and Gaussian-blending technique. Specifically, the image is first converted to a numpy array, inverted, then blurred, and finally blended back with the original input using pixel-wise operations. This step produces a sketch-like image that preserves structural edges.

	\paragraph{Random Channel Mixing.}
	The core transformation is implemented in \texttt{Fuse\_RGB\_Gray\_Sketch}, where the input image is first split into its red ($r$), green ($g$), and blue ($b$) channels. A grayscale image is also computed from the original input. Based on a randomly drawn probability $p \in [0, 1]$, one of four transformation cases is selected:
	
	\begin{itemize}
		\item If $p < G$: convert the image to pure grayscale by repeating the single-channel grayscale three times to form an RGB image.
		\item If $G \leq p < G + G_{rgb}$: mix the RGB channels with grayscale using random channel positions. 
		\item If $G + G_{rgb} \leq p < G + G_{rgb} + S_{rgb}$: similarly, mix RGB channels with a sketch image to simulate sketch-style corruption.
		\item Otherwise: return the original RGB image unchanged.
	\end{itemize}

	\paragraph{Channel Fusion Strategy.}
	The \texttt{random\_choose} function constructs a mixed image by randomly shuffling and selecting three channels from the list $[r, g, b, gray\_or\_sketch, gray\_or\_sketch]$. This ensures that at least one or two channels are from the distorted source, while preserving RGB compatibility. The repetition of the grayscale/sketch entry ensures higher probability of being included in the output.

	\section{Analysis of Adversarial Defense}
	To further validate the practical value of our proposed method, we integrate it into the adversarial training pipeline to investigate its efficacy in enhancing model robustness against multimodal attacks. Adversarial training, a widely used defense mechanism, strengthens model resilience by incorporating adversarial samples during the training process. This section explores whether adversarial training using adversarial samples generated by the proposed Multiform Attack can significantly improve defense performance across heterogeneous modalities.

	Our device is equipped with two NVIDIA A40 GPUs, each with 48GB of memory.
	We conducted experiments on the RegDB dataset to evaluate the baseline model AGW. Adversarial samples were generated using our Gradient-Evolutionary Multiform Optimization framework with $\epsilon=8$. During adversarial training, each mini-batch consisted of 50\% adversarial samples and 50\% original samples. We compared our method with the state-of-the-art cross-modal attack method MUA to test the model's adversarial robustness. Performance was assessed using Rank-1 accuracy and mAP to evaluate the robustness of the trained models.
	
	\begin{table}[ht]
		\setlength{\tabcolsep}{2.5pt} % 调整列间距
		\centering
		\caption{Defense performance comparison between normal training and adversarial training using MUA attack and our attack. Experiments were conducted on the RegDB dataset using AGW as the baseline. Rank-1 and mAP are reported.}
		\begin{tabular}{l|cc|cc}
			\hline	
			\hline
			\textbf{Training Type} & \multicolumn{2}{c|}{\textbf{MUA attack}} & \multicolumn{2}{c}{\textbf{Our attack}} \\
			& Rank-1 (\%) & mAP (\%) & Rank-1 (\%) & mAP (\%) \\
			\hline
			Normal Training & 1.85 & 3.21 & 0.83 & 1.08 \\
			Adv. Training (MUA attack) & 34.53 & 30.31 & 27.14 & 24.36 \\
			Adv. Training (Our attack) & 40.37 & 35.22 & 38.35 & 33.41 \\
			\hline
			\hline
		\end{tabular}
		\label{defense_results}
	\end{table}

	Tab.~\ref{defense_results} presents the experimental results. 
	From the table, we observe that models trained with normal training exhibit poor robustness to both MUA and our proposed attack, achieving only 1.85\% Rank-1 accuracy and 3.21\% mAP under MUA attack, and even lower results of 0.83\% Rank-1 accuracy and 1.08\% mAP under our attack. 
	These results highlight the vulnerability of the baseline model to cross-modal adversarial attacks, especially under the stronger perturbations generated by our method.
	
	When adversarial training is applied using MUA attack samples, the model's robustness improves significantly, achieving 34.53\% Rank-1 accuracy and 30.31\% mAP under MUA attacks. 
	However, under our attack, the performance drops to 27.14\% Rank-1 accuracy and 24.36\% mAP. 
	This demonstrates that adversarial training with MUA samples enhances robustness against MUA-specific attacks but shows limited generalization to more transferable attacks, such as those generated by our method.
	
	Adversarial training using samples generated by our method further improves robustness across both attack types. 
	The model achieves 40.37\% Rank-1 accuracy and 35.22\% mAP under MUA attacks, demonstrating improved generalization across adversarial strategies. 
	Similarly, the model achieves 38.35\% Rank-1 accuracy and 33.41\% mAP under our own attack, indicating that adversarial training with our generated samples better prepares the model against diverse and strong cross-modal perturbations.
	
	These results highlight the importance of adversarial training in enhancing cross-modal robustness. 
	Furthermore, adversarial samples generated by our gradient-evolutionary multiform optimization framework can serve as effective training examples for improving adversarial robustness. 
	This suggests that MA is useful not only for robustness evaluation but also for constructing stronger adversarial training data in cross-modal ReID systems.

	\section{Experimental Access Protocol and Threat Model}
	\label{sec:threat_model}
	
	This section describes the attacker access, data usage, model access, and evaluation protocol used in our experiments. 
	Our goal is to avoid ambiguity between the restricted pairwise attack setting used for fair comparison on standard cross-modality ReID benchmarks and the heterogeneous transfer setting used to evaluate cross-dataset and cross-modality generalization.

	We use two evaluation protocols:
	
	\paragraph{Protocol I: Restricted compatible setting.}
	This protocol is used for the standard benchmark comparisons in Tab.~1 and Tab.~2. 
	The purpose is to compare MA with existing attack methods under the same modality-pair setting. 
	In this setting, MA does not use additional external datasets, additional modalities, or an enlarged surrogate-model ensemble beyond the evaluated source modality pair. 
	The perturbation is optimized only using the training data and model access available under the same setting as the compared baselines. 
	All methods are evaluated under the same perturbation budget, query-gallery protocol, data split, and modality-pair setting.
	
	\paragraph{Protocol II: Heterogeneous transfer setting.}
	This protocol is used for the cross-dataset and cross-modality transfer experiments in Tab.~3. 
	The purpose is to evaluate whether a perturbation optimized on accessible source domains can transfer to a held-out target domain. 
	In this setting, the attacker may use multiple source or auxiliary domains during optimization, but the target domain is strictly excluded from perturbation generation, model selection, and hyperparameter tuning. 
	For fairness, all compared methods, including CMPS and MUA, are optimized using the same source and non-target auxiliary domains. 
	The held-out target dataset/model is used only once for final evaluation.

	Table~\ref{tab:threat_model_summary} summarizes the attacker access used in each experimental setting. 
	``Source dataset'' denotes the dataset on which the main universal perturbation is optimized. 
	``Auxiliary datasets'' denote additional accessible non-target datasets used only in the heterogeneous transfer setting. 
	``Source/surrogate models'' denote models available to the attacker for gradient optimization and evolutionary fitness evaluation. 
	``Target model'' denotes the model used only for final evaluation.

	\begin{table*}[t]
		\centering
		\caption{Attacker access and evaluation protocol used in different experiments. 
			The target dataset/model is never used for perturbation optimization or hyperparameter tuning.}
		\label{tab:threat_model_summary}
		\resizebox{\textwidth}{!}{
			\begin{tabular}{lcccccc}
				\toprule
				Experiment 
				& Protocol 
				& Source dataset(s) 
				& Auxiliary dataset(s) 
				& Source / surrogate model access 
				& Target dataset / model 
				& Label access \\
				\midrule
				RegDB V$\rightarrow$T / T$\rightarrow$V 
				& Restricted 
				& RegDB 
				& None 
				& AGW on RegDB 
				& RegDB / AGW 
				& Source labels only \\
				
				LLCM V$\rightarrow$I / I$\rightarrow$V 
				& Restricted 
				& LLCM 
				& None 
				& DEEN on LLCM 
				& LLCM / DEEN 
				& Source labels only \\
				
				SYSU $\rightarrow$ CnMix 
				& Heterogeneous transfer 
				& SYSU 
				& RegDB, Sketch 
				& Models trained on accessible source/auxiliary domains 
				& CnMix / held-out CnMix model 
				& Source/auxiliary labels only \\
				
				SYSU $\rightarrow$ Sketch 
				& Heterogeneous transfer 
				& SYSU 
				& RegDB, CnMix 
				& Models trained on accessible source/auxiliary domains 
				& Sketch / held-out Sketch model 
				& Source/auxiliary labels only \\
				
				CnMix $\rightarrow$ SYSU 
				& Heterogeneous transfer 
				& CnMix 
				& RegDB, Sketch 
				& Models trained on accessible source/auxiliary domains 
				& SYSU / held-out SYSU model 
				& Source/auxiliary labels only \\
				
				Sketch $\rightarrow$ SYSU 
				& Heterogeneous transfer 
				& Sketch 
				& RegDB, CnMix 
				& Models trained on accessible source/auxiliary domains 
				& SYSU / held-out SYSU model 
				& Source/auxiliary labels only \\
				\bottomrule
		\end{tabular}}
	\end{table*}
	
	\section{Leave-One-Modality-Out Evaluation for Unseen-Modality Transfer}
	\label{sec:lomo_protocol}
	
	This section clarifies how we evaluate transferability to unseen modalities and how this evaluation differs from the standard pairwise cross-modality attack setting. 
	The purpose is to avoid ambiguity between two related but different questions: 
	(i) whether an attack is effective under a known cross-modality query-gallery protocol, and 
	(ii) whether an attack optimized on accessible source modalities can generalize to a held-out target modality or target domain that is not used during perturbation generation.
	
	\subsection{Known-Pair Cross-Modality Attack vs. Unseen-Modality Transfer}
	
	We distinguish between two evaluation settings:
	
	\paragraph{Known-pair cross-modality attack.}
	This setting corresponds to the standard benchmark comparisons on RegDB and LLCM. 
	For example, RegDB Visible$\rightarrow$Thermal and Thermal$\rightarrow$Visible evaluate whether the adversarial perturbation can attack a known visible-thermal ReID protocol. 
	Similarly, LLCM Visible$\rightarrow$Infrared and Infrared$\rightarrow$Visible evaluate attack effectiveness under a known visible-infrared protocol. 
	These experiments are used to compare MA with existing attacks under the same benchmark, perturbation budget, query-gallery protocol, data split, and model setting.

	\paragraph{Unseen-modality or held-out-domain transfer.}
	This setting evaluates whether a perturbation optimized on accessible source modalities can transfer to a target modality or target domain that is excluded from all stages of perturbation generation. 
	A target modality/domain is considered \emph{unseen} if its training data, validation data, test data, model parameters, and retrieval feedback are not used during universal perturbation learning, sparse evolutionary refinement, model selection, or hyperparameter tuning. 
	Only after the perturbation is fixed do we evaluate it on the held-out target query-gallery protocol.
	
	\subsection{Leave-One-Modality-Out Protocol}
	
	To make the unseen-modality claim explicit, we adopt a leave-one-modality-out protocol. 
	Let $\mathcal{D} = \{D_1, D_2, \ldots, D_M\}$ denote the set of available modality/domain datasets, such as visible-infrared, visible-thermal, low-light infrared, sketch, and channel-mixed domains. 
	For each target domain $D_t$, we define:
	\begin{equation}
		\mathcal{D}_{\mathrm{src}} = \mathcal{D} \setminus \{D_t\}.
	\end{equation}
	The perturbation is optimized only on $\mathcal{D}_{\mathrm{src}}$, while $D_t$ is held out completely until final evaluation. 
	The final perturbation $\Delta$ is then directly applied to query images from $D_t$, and gallery images remain clean.
	
	The leave-one-modality-out protocol follows three rules:
	
	\begin{itemize}
		\item \textbf{No target data during optimization.} 
		Images, labels, feature statistics, cluster centroids, covariance matrices, and attention maps from the held-out target domain are not used during either the gradient-based universal perturbation stage or the evolutionary sparse refinement stage.
		
		\item \textbf{No target model access.} 
		The target ReID model is not used as a surrogate model and is not queried during perturbation generation. 
		Fitness evaluation in the evolutionary stage is computed only on accessible source or auxiliary surrogate models.
		
		\item \textbf{No target validation feedback.} 
		Hyperparameters, including the perturbation budget, population size, number of generations, sparsity budget, and attention-filtering threshold, are selected using source-domain validation only. 
		The target domain is used only once for final evaluation.
	\end{itemize}
	
	\subsection{Auxiliary Domains and Target Exclusion}
	
	In heterogeneous transfer experiments, MA may use multiple accessible non-target domains as auxiliary source domains. 
	This does not violate the unseen-modality protocol as long as the final target domain is excluded from all optimization and validation steps. 
	For fairness, when auxiliary domains are used, all compared methods are given the same source and non-target auxiliary domains. 
	Thus, improvements cannot be attributed to additional target-domain access.
	For example, in a transfer setting denoted as SYSU$\rightarrow$CnMix, SYSU is the primary source domain and CnMix is the held-out target domain. 
	Other non-target datasets may be used as auxiliary source domains, but CnMix images, CnMix labels, CnMix feature statistics, CnMix-trained models, and CnMix validation results are not used before final testing. 
	The same source and auxiliary domains are used for CMPS, MUA, and MA.

	In the restricted compatible setting, the terms ``source'' and ``target'' refer to the two modalities within the same benchmark protocol, such as visible-to-thermal and thermal-to-visible on RegDB. 
	No additional source domain is introduced in this setting. 
	This setting is designed to ensure direct comparability with prior cross-modality ReID attacks.
	In the heterogeneous transfer setting, the target domain denotes the held-out dataset/model on which the final attack performance is reported. 
	The target domain is not used during universal perturbation learning, sparse evolutionary refinement, model selection, or hyperparameter tuning. 
	Auxiliary datasets, when used, are limited to non-target training splits and are shared across all compared methods. 
	Thus, performance improvements in this setting cannot be attributed to target-domain leakage or target validation feedback.

	For all experiments, adversarial perturbations are applied only to query images. 
	Gallery images remain clean and unchanged. 
	The query-gallery splits follow the standard evaluation protocol of each benchmark. 
	For single-channel modalities such as thermal or infrared images, the input is replicated to three channels to match the standard input format of the ReID backbones. 
	The same preprocessing rule is applied to all compared methods.

	Hyperparameters are selected using only source-domain training or validation data. 
	For Protocol I, hyperparameters are tuned on the validation split of the corresponding benchmark source setting without using test-set results. 
	For Protocol II, hyperparameters are selected using only accessible source and auxiliary domains. 
	No target-domain validation set, target test labels, or repeated target feedback is used for hyperparameter tuning.
	When evaluating transferability in Protocol II, all compared methods are given the same accessible source and auxiliary domains. 
	That is, if MA uses a set of accessible non-target domains for optimization, CMPS and MUA are also optimized using the same accessible domains under their respective optimization procedures. 
	Therefore, the comparison isolates the effect of the perturbation optimization strategy rather than differences in data access.

	\subsection{Fairness Analysis of Multi-Source Optimization}
	\label{sec:fairness}
	
	To disentangle the contribution of the algorithmic design from the benefit of additional source data, we conduct an ablation experiment. We introduce a restricted variant of MA, denoted MA$_{\text{single}}$, which uses only the primary source domain pair and discards all auxiliary domains. The comparison between MA$_{\text{single}}$ and MA isolates the effect of auxiliary source domains.
	
	Table~\ref{tab:ablation_multi_source} reports the results on the SYSU $\rightarrow$ CnMix transfer task. Several observations can be made:
	\begin{itemize}
	\item Adding auxiliary source domains further improves MA's performance. This confirms that the evolutionary refinement stage can effectively leverage heterogeneous source data when such data are available.
	\item The performance of MA does not solely rely on extra data access. The algorithmic design already achieves competitive performance in the restricted single-source setting, while auxiliary domains provide additional improvements.
	\end{itemize}

	\begin{table}[h]
		\centering
		\caption{Ablation study on the effect of auxiliary source domains for the SYSU $\rightarrow$ CnMix transfer task. $\mathrm{MA}_{\mathrm{single}}$ is trained with only the primary source domain pair and receives no additional auxiliary data, whereas full MA additionally leverages accessible auxiliary domains, i.e., RegDB and Sketch. All methods are compared under the same perturbation budget $\epsilon=8$. Lower Rank-1 and mAP indicate stronger attack performance.}
		\label{tab:ablation_multi_source}
		\begin{tabular}{lcc}
			\hline
			Method & Rank-1 (\%) & mAP (\%) \\
			\hline
			MA$_{\text{single}}$  & 8.87 $\pm$ 0.25 & 8.13 $\pm$ 0.19 \\
			MA  & \textbf{6.42} $\pm$ 0.19 & \textbf{5.74} $\pm$ 0.14 \\
			\hline
		\end{tabular}
	\end{table}
	
	\begin{figure*}[ht]
		\centering
		\includegraphics[width=1\linewidth]{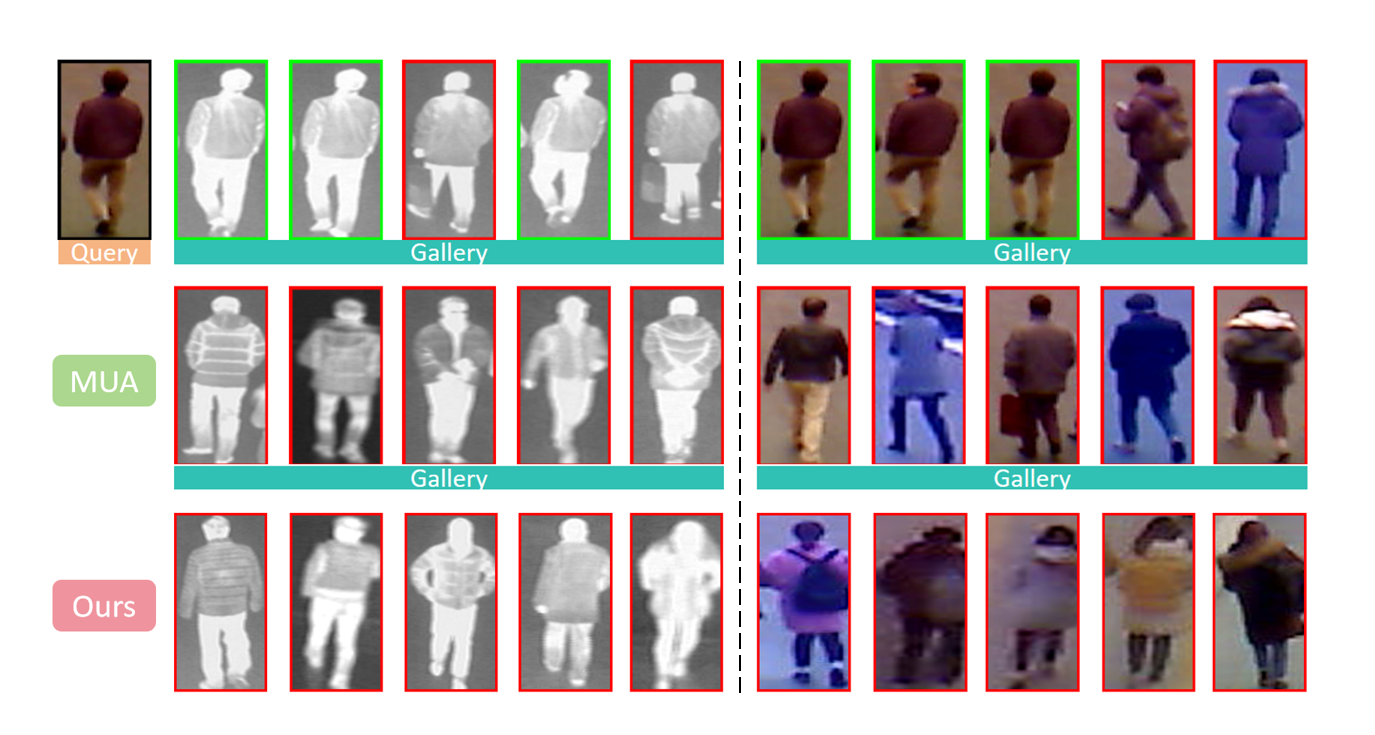}
		\caption{Comparison with the state-of-the-art attack method MUA attack retrieval results. Before and after our attack on RegDB, the top five predictions of DDAG. Green boxes indicate correctly matched images, while red boxes indicate mismatched images.} 
		\label{reid}
	\end{figure*}

	\section{Qualitative Comparison with State-of-the-Art Attack Methods}
	
	To further demonstrate the effectiveness of our proposed attack, we provide a qualitative comparison with the state-of-the-art MUA method in Fig.~\ref{reid}. The visualization shows the top-5 retrieval results produced by the DDAG baseline on the RegDB dataset. For each query (leftmost column), we present the retrieved gallery images under three different conditions: before attack (top row), after MUA attack (middle row), and after our proposed attack (bottom row). Green boxes denote correctly matched identities, while red boxes indicate mismatched identities.
	
	The right panel illustrates the retrieval results in the visible-to-visible direction. Interestingly, we observe that our attack not only causes complete mismatches but also leads to more diverse false positives among the top-5 retrieval results. In contrast, the MUA attack tends to yield semantically similar but incorrect matches, whereas our method produces more scattered and unrelated retrievals. This suggests that the feature space has been more thoroughly disrupted, indicating that our attack interferes with the learned metric at a more fundamental level, rather than merely adding noise near class boundaries.
	
	These visual results confirm that our multiform adversarial framework produces more deceptive and disruptive perturbations compared to MUA. By leveraging both gradient guidance and structure-aware evolutionary search, our method introduces broader misalignment in the learned feature manifold, achieving stronger attack effectiveness even under high-confidence retrieval scenarios.

	\begin{figure*}[]
		\centering
		\includegraphics[width=1\linewidth]{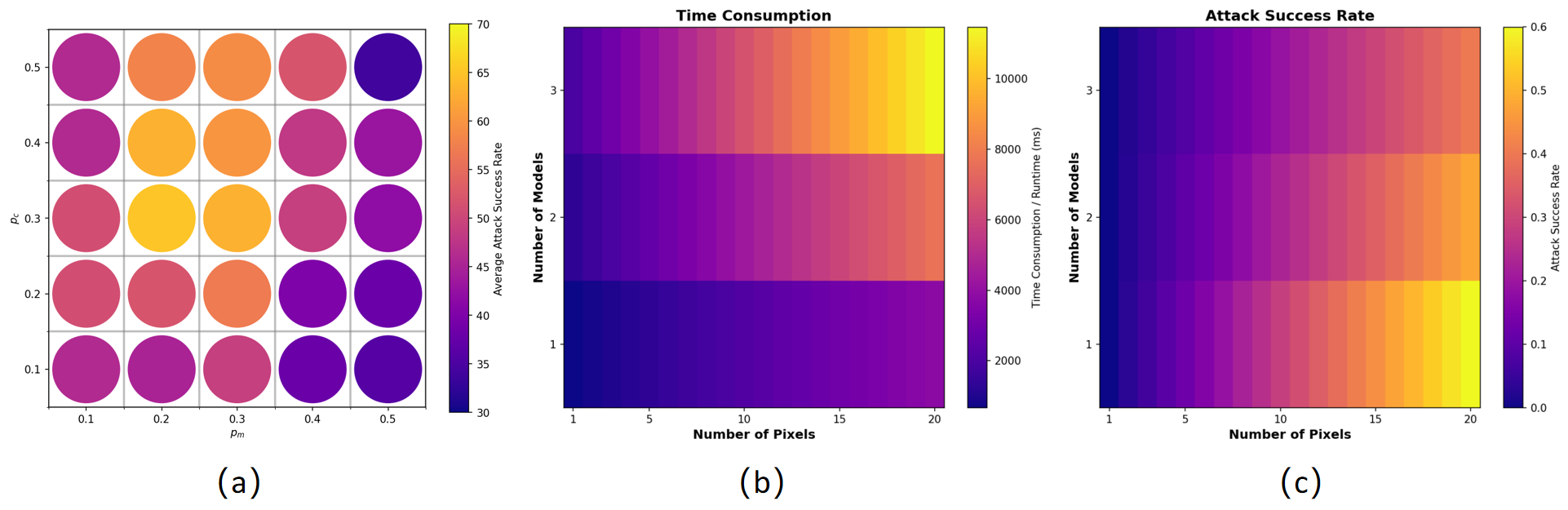}
		%\setlength{\abovecaptionskip}{-0.2cm}
		%\captionsetup{skip=0pt} % 设置 skip 参数为5pt（可以根据需要调整距离）
		\caption{
			(a) The impact of mutation rates (\(p_m\)) and crossover rates (\(p_c\)) on the attack success rate during evolutionary search. 
			(b) The relationship among the number of perturbed pixels, the number of modalities, and time consumption. 
			(c) The interplay among the number of perturbed pixels, the number of modalities, and the attack success rate.} 
		\label{xxxx1}
	\end{figure*}

	\section{Ablation Experiment}
	\subsection{Ablation Study on Evolutionary Search Factors}
	Fig.\ref{xxxx1} (a), (b), and (c) illustrate the impact of various factors on attack performance during evolutionary search.
	
	Fig.\ref{xxxx1}(a) shows the impact of different crossover rates ($p_c$) and mutation rates ($p_m$) on the attack success rate when using evolutionary search alone. The heatmap shows that higher crossover rates ($p_c$) generally improve success rates, especially when combined with moderate mutation rates ($p_m$). It can be observed that the combination of $p_m = 0.2$ and $p_c = 0.3$ achieves the highest success rate. Unless otherwise specified, we adopt this parameter combination in our experiments.
	
	Fig.\ref{xxxx1}(b) demonstrates the relationship between the number of perturbed pixels, the number of modalities, and time consumption. Key observations include: (1) The attack success rate improves as the number of perturbed pixels increases; (2) Time consumption increases proportionally with the number of perturbed pixels; (3) Time consumption further rises with the number of modalities due to additional computational overhead; (4) The attack success rate declines as the number of modalities increases, highlighting the challenges of optimizing across diverse modalities.
	
	Fig.\ref{xxxx1}(c) illustrates the interplay between the number of perturbed pixels, the number of modalities, time consumption, and attack success rate. While increasing the number of perturbed pixels improves the attack success rate, it also significantly increases computational cost, particularly in multi-modality scenarios.

	\begin{figure}[h]
		\centering
		\includegraphics[width=0.8\linewidth]{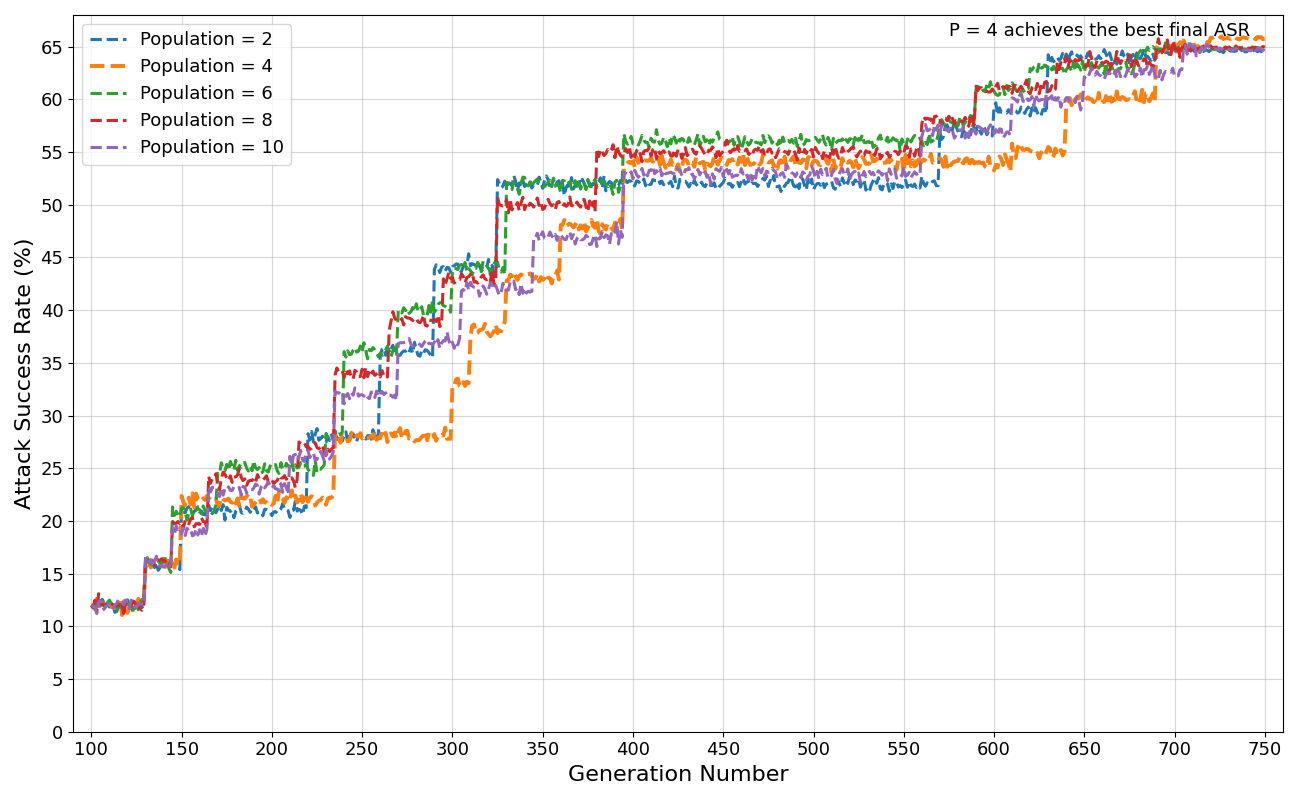}
		\caption{Attack success rate by generation number for different population sizes.} 
		\label{XXX2}
	\end{figure}
	
	Fig.~\ref{XXX2} illustrates the sensitivity of the evolutionary refinement stage to the generation number and population size. 
	Among the tested settings, $P=4$ provides the best trade-off between attack success rate and computational efficiency: it achieves competitive or slightly better final attack success while avoiding the additional cost of larger populations. 
	Therefore, we adopt $G=200$ and $P=4$ as the default configuration in all main experiments.
	
	\begin{figure}[]
		\centering
		\includegraphics[width=0.75\linewidth]{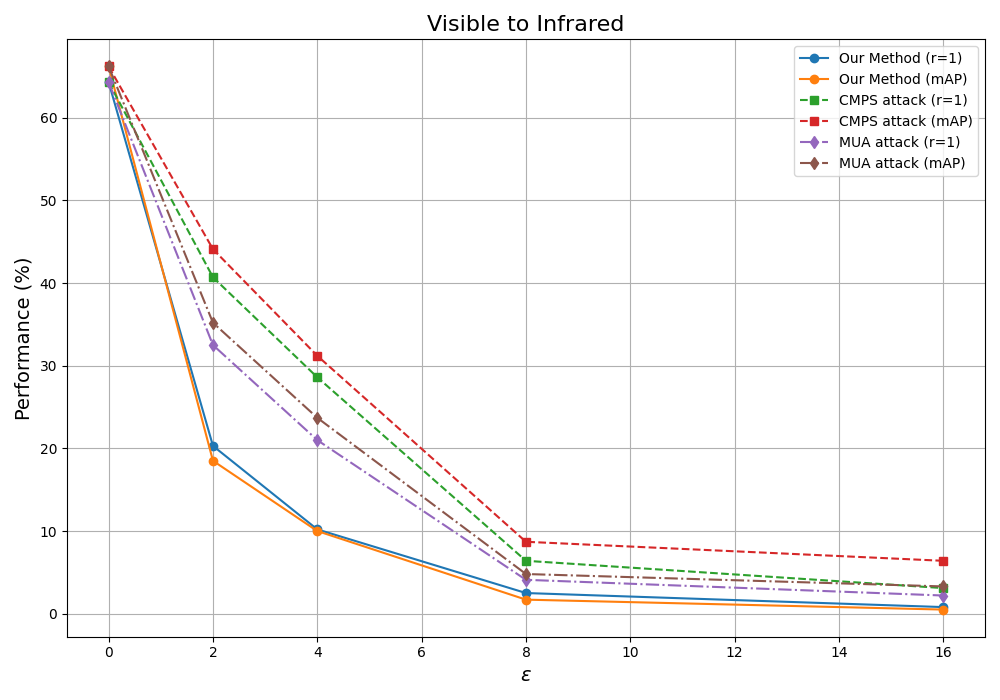}
		%\setlength{\abovecaptionskip}{-0.2cm}
		%\captionsetup{skip=0pt} % 设置 skip 参数为5pt（可以根据需要调整距离）
		\caption{
			Performance degradation under different perturbation strengths (\(\epsilon\)) on the RegDB dataset (Visible to Thermal). We compare our proposed method with CMPS and MUA attacks using Rank-1 accuracy and mAP as evaluation metrics.Note that lower values of Rank-1 and mAP indicate stronger attack effectiveness, as they reflect greater disruption to the model’s retrieval performance. 
		}
		\label{xxxx2}
	\end{figure}
	
	\subsection{Experimental Comparison under Varying Perturbation Strengths}
	\label{sec:ablation_strength}
	
	To investigate the robustness of cross-modal ReID models under different perturbation magnitudes, we conduct an ablation study varying the \(\ell_\infty\) perturbation budget \(\epsilon\) from 0 to 16. The results on the RegDB dataset (visible-to-infrared direction) are presented in Fig.~\ref{xxxx2}, comparing our method with the CMPS and MUA attacks in terms of Rank-1 accuracy and mAP.
	
	All methods exhibit declining performance as \(\epsilon\) increases, indicating that larger perturbations lead to stronger degradation of model predictions. However, the rate of performance drop differs across methods, reflecting their varying effectiveness in adversarial generation. Our method consistently leads to the most significant performance degradation across all \(\epsilon\) values. For example, at \(\epsilon=4\), our method reduces Rank-1 accuracy to only 10.2\% compared to 21.3\% by MUA and 28.6\% by CMPS, showing a stronger attack capability under the same constraint. Even at \(\epsilon=2\), our method is already able to suppress the model significantly, unlike CMPS which maintains relatively high performance.
	
	Compared to CMPS, the MUA attack yields stronger degradation at every perturbation level, suggesting better transferability and optimization. Nevertheless, it remains weaker than our method, highlighting the benefit of incorporating multiform gradient evolution and structure-aware search in crafting more disruptive perturbations. The empirical results validate the superiority of our proposed attack across a wide range of perturbation budgets. It confirms the benefit of combining gradient-based updates with manifold-aligned evolutionary strategies for generating highly disruptive adversarial examples in cross-modal ReID.

	\begin{table}[t]
		\centering
		\caption{
			Cross-model transferability of our attack under different perturbation strengths \(\epsilon\). 
			We report Rank-1 accuracy and mAP (\%) when perturbations are crafted on a surrogate model 
			and transferred to a different target model. Lower values indicate stronger attack effectiveness.
		}
		\setlength{\tabcolsep}{4pt}
		\begin{tabular}{c|cc|cc|cc}
			\hline
			\multirow{2}{*}{\(\epsilon\)} & 
			\multicolumn{2}{c|}{ResNet \(\rightarrow\) Swin} & 
			\multicolumn{2}{c|}{Swin \(\rightarrow\) VGG} & 
			\multicolumn{2}{c}{VGG \(\rightarrow\) ResNet} \\
			& Rank-1 & mAP & Rank-1 & mAP & Rank-1 & mAP \\
			\hline
			2  & 42.9 & 39.1 & 35.7 & 32.4 & 38.2 & 35.6 \\
			4  & 30.6 & 26.3 & 20.3 & 18.1 & 26.1 & 23.2 \\
			8  & 10.7 & 9.1 & 7.0  & 6.3  & 12.4 & 10.5  \\
			16 & 6.7  & 5.2  & 3.1  & 2.5  & 5.3  & 4.2  \\
			\hline
		\end{tabular}
		\label{tab:transfer_eps}
	\end{table}

	\subsection{Cross-Model Transferability under Different Perturbation Strengths}
	\label{sec:transferability}
	
	To further investigate how different network architectures respond to adversarial perturbations, we evaluate the cross-model transferability of our method under increasing perturbation strengths \(\epsilon\). As shown in Tab.~\ref{tab:transfer_eps}, we consider three representative transfer directions: ResNet50 \(\rightarrow\) Swin, Swin \(\rightarrow\) VGG, and VGG \(\rightarrow\) ResNet50, where adversarial perturbations are generated on a source (surrogate) model and applied to a different target model.
	
	Several noteworthy patterns emerge from the results:
	Swin exhibits the strongest robustness as a target model. When perturbations crafted on ResNet50 are transferred to Swin, the drops in Rank-1 accuracy and mAP are relatively mild. This suggests that the hierarchical attention mechanism in Swin provides enhanced resistance against transferred adversarial noise.
	Perturbations generated on Swin transfer most effectively to VGG. Among the three transfer settings, the Swin \(\rightarrow\) VGG direction consistently yields the most severe performance degradation. This indicates that Transformer-based architectures may produce perturbations aligned with higher-level semantic features, thus improving their transferability across architectures.
	VGG-originated perturbations are relatively weaker but remain harmful. Although adversarial examples generated on VGG appear less aggressive overall, they still substantially reduce the performance of ResNet50. 
	
	In summary, these findings confirm that the perturbations produced by our method are not only effective within a single model but also exhibit strong and interpretable transferability across fundamentally different backbone architectures. The consistent decrease in both Rank-1 and mAP across all configurations further validates the generality and effectiveness of our approach.

	\section{Discussion}
	We recognize the positive value of adversarial attack research. It reveals vulnerabilities in existing systems, prompting academia and industry to make in-depth improvements to the robustness of machine learning models. The positive impact of this study lies in its potential to combine adversarial training with the attack methods presented to enhance system security and bring positive social impacts. Therefore, we emphasize the importance of conducting adversarial attack research within an ethical framework and encourage further development of defensive technologies to build a safer and more reliable technological environment.

	%%%%%%%%%%%%%%%%%%%%%%%%%%%%%%%%%%%%%%%%%%%%%%%%%%%%%%%%%%%%
	
	%\input{checklist.tex}

\end{document}